\documentclass{article}

\usepackage{microtype}
\usepackage{graphicx}
\usepackage{booktabs} 
\usepackage{amsmath}
\usepackage[T1]{fontenc}
\usepackage{hyperref}
\usepackage{subfig}

\newcommand{\D}{\mathcal{D}}

\renewcommand{\P}{\mathbb{P}}

\newcommand{\norm}[1]{\left\| #1 \right\|}

\DeclareMathOperator{\Tr}{Tr}
\newcommand{\E}{\mathbb{E}}

 \usepackage[accepted]{icml2024}

\usepackage{amsmath}
\usepackage{amssymb}
\usepackage{mathtools}
\usepackage{amsthm}

\usepackage[capitalize,noabbrev]{cleveref}

\theoremstyle{plain}
\newtheorem{theorem}{Theorem}[section]

\theoremstyle{definition}

\theoremstyle{remark}

\usepackage[textsize=tiny]{todonotes}

\icmltitlerunning{Understanding the Learning Dynamics of Alignment with Human Feedback}

\begin{document}

\twocolumn[
\icmltitle{Understanding the Learning Dynamics of Alignment with Human Feedback}



\icmlsetsymbol{equal}{*}

\begin{icmlauthorlist}
\icmlauthor{Shawn Im}{yyy}
\icmlauthor{Yixuan Li}{yyy}
\end{icmlauthorlist}

\icmlaffiliation{yyy}{Department of Computer Sciences, University of Wisconsin-Madison}

\icmlcorrespondingauthor{Shawn Im}{shawnim@cs.wisc.edu}
\icmlcorrespondingauthor{Yixuan Li}{sharonli@cs.wisc.edu}

\icmlkeywords{Machine Learning, ICML}

\vskip 0.3in
]



\printAffiliationsAndNotice{ }  

\begin{abstract}
Aligning large language models (LLMs) with human intentions has become a critical task for safely deploying models in real-world systems. While existing alignment approaches have seen empirical success, theoretically understanding how these methods affect model behavior remains an open question. Our work provides an initial attempt to theoretically analyze the learning dynamics of human preference alignment. We formally show how the distribution of preference datasets influences the rate of model updates and provide rigorous guarantees on the training accuracy. Our theory also reveals an intricate phenomenon where the optimization is prone to prioritizing certain behaviors with higher preference distinguishability. We empirically validate our findings on contemporary LLMs and alignment tasks, reinforcing our theoretical insights and shedding light on considerations for future alignment approaches. 
\textcolor{red}{\emph{Disclaimer: This paper contains potentially offensive text; reader discretion is advised.}}
\end{abstract}

\section{Introduction}
\label{submission}

Large language models (LLMs) have demonstrated remarkable abilities to generate human-like text and acquire diverse capabilities~\citep{brown2020language, wei2022emergent, anil2023palm}. However, these models are not necessarily aligned with human preferences and can inadvertently produce harmful or undesirable outputs. Thus, aligning language models with human preferences has become an important problem, which ensures that these models exhibit safe and desirable behavior. Existing alignment approaches share the basis of reinforcement learning from human preferences (RLHF)~\citep{christiano2017deep, ziegler2019finetuning, ouyang2022training, bai2022training}, which involves fitting a reward model to the preference data and optimizing a language model policy for high reward through reinforcement learning. Despite the empirical success and wide adoption in real-world systems~\citep{openai2023gpt4, anthropic2023claude, touvron2023llama}, theoretical understanding of alignment with human preferences is still in its infancy.

In particular, analyzing the learning dynamics of RLHF theoretically is a challenging task, as it requires understanding both the learned reward model and how it guides the policy learned during reinforcement learning. Moreover, the computational expense associated with RLHF, involving multiple models, adds to the complexity.  Recently, 
a reparameterization of RLHF called Direct Preference Optimization (DPO) \citep{rafailov2023direct} has emerged as a promising alternative, which directly optimizes the policy to best satisfy preferences and circumvents the need for RL training. 
\citet{rafailov2023direct} showed that under mild assumptions, the optimal policy under the DPO objective is the same as the optimal policy using RLHF. The equivalence makes rigorously analyzing how models change when learning human preferences more tractable. With DPO, it is sufficient to consider the relationship between the policy and the dataset.

In this paper, we provide a theoretical analysis of how DPO dynamics change based on the distributional properties of the preference dataset. We characterize the data distributions through the lens of \emph{preference distinguishability}, which refers to how far apart the distributions for the preferred and non-preferred responses are. Based on this notion, we provide learning guarantees on how preference distinguishability impacts the rate of weight parameter updates under the DPO objective (Theorem~\ref{thm:norm}), along with a lower bound for the accuracy (Theorem~\ref{thm:cosine} and Theorem~\ref{thm:boundary}).  
Our theorem indicates that, under the same training configuration, higher distinguishability leads to a faster rate of change in weight parameters and a more rapid decrease of loss. Our theoretical insight has practical implications for alignment training on diverse preference datasets encompassing various topics and behaviors of differing distinguishability. In particular, we reveal an intricate prioritization effect, where DPO is prone to prioritize learning behaviors with higher distinguishability and as a result, may deprioritize the less distinguishable yet crucial ones. Such an effect can manifest in real systems, where for example, certain political views or ideological beliefs may be prioritized in the learning process over others.

We empirically validate our theoretical insights and show that they generalize to practical LLMs. Leveraging the latest Llama-2 model~\cite{touvron2023llama}, we conduct extensive experiments by training on diverse preference datasets using the DPO objective.  Consistent with our theory, our results indicate that behaviors with higher distinguishability exhibit a more rapid rate of loss reduction.  Moreover, when training multiple behaviors simultaneously, the effect of prioritization remains influential in the practical setting.  Notably, we observe that models trained with DPO are more susceptible to being unaligned or misaligned compared to their corresponding base models.  These findings shed light on the vulnerability of RLHF and DPO-trained models, and underscore the importance of considering preference or behavior prioritization in alignment training.

We summarize our key contributions in the following:
\begin{itemize}
\item To the best of our knowledge, we provide a first attempt to understand the learning dynamics of the alignment approach from a rigorous theoretical point of view.
  \item We provide new learning guarantees on how preference distinguishability impacts the rate of weight parameter updates under the DPO objective (Theorem~\ref{thm:norm}), along with a lower bound on training accuracy (Theorem~\ref{thm:cosine} and Theorem~\ref{thm:boundary}).  
  \item We empirically validate our findings on modern LLMs and preference datasets containing diverse behaviors, reinforcing our theoretical insights and inspiring future research on practical algorithms for alignment.
\end{itemize}

\section{Preliminaries}

\paragraph{Notations.} We denote $\pi_{\theta}$ as a language model policy parameterized by $\theta$, which takes in an input prompt $x$, and outputs a discrete probability distribution $\pi_\theta(\cdot|x)$ over the  vocabulary space $\mathcal{V}$. 
$\pi_\theta(y | x)$ refers to the model's probability of outputting response $y$ given input prompt $x$. 
Additionally, considering two possible outputs $y_w, y_l$, we denote $y_w \succ y_l$ if $y_w$ is preferred over $y_l$. We call $y_w$ the preferred response and $y_l$ the less preferred response.

\paragraph{RLHF Overview.} 
Reinforcement Learning from Human Feedback (RLHF) is a widely used paradigm for learning desirable behaviors based on human preferences~\citep{christiano2017deep, ziegler2019fine, ouyang2022training, bai2022training}. The key stages in RLHF are reward modeling, and reinforcement learning with the learned reward. Here we provide a brief recap of the two stages, respectively. 

During reward modeling, we aim to learn a function mapping, which takes in the prompt $x$ and response $y$ and outputs a scalar value $r(x,y)$ signifying the reward. A preferred response should receive a higher reward, and vice versa. Under the Bradley-Terry model~\cite{bradley1952rank}, the preference distribution is modeled as 
\begin{equation}
    \label{eq:bt-model}
    p^*(y_w \succ y_l | x) = \sigma(r^*(x, y_w) - r^*(x, y_l)),
\end{equation}
where $\sigma$ is the sigmoid function. Given the empirical dataset $\mathcal{D}=\{(x_i, y_{w,i}, y_{l,i})\}_{i=1}^n$ sampled from the preference distribution $p^*$,  we can learn the reward function via maximum likelihood estimation, which is equivalent to optimizing the following binary classification objective:
\begin{equation}
    \mathcal{L}_R = -\mathbb{E}_{(x,y_w,y_l)\sim \mathcal{D}}[\log \sigma(r(x,y_w) - r(x,y_l))].
\end{equation}

Using the learned reward function, the model is fine-tuned with reinforcement learning to maximize the following objective
\begin{equation}
    \label{eq:rlhf}
    R(\pi_\theta) = \E_{\pi_\theta} [r(x, \hat{y})] - \beta \log \frac{\pi_\theta(\hat{y} | x)}{\pi_{\text{ref}}(\hat{y} | x)},
\end{equation}
where $\hat{y}$ is the output generated by the current model's policy $\pi_\theta$ for the prompt $x$, $\pi_{\text{ref}}$ is the policy of the model before any steps of RLHF, and $\beta$ is a hyperparameter. We can view this objective as maximizing the expected reward with KL regularization weighted by $\beta$. 

\paragraph{Direct Preference Optimization.} 
Analyzing the dynamics of RLHF rigorously is a difficult task as it requires understanding both the learned reward model and how it guides the policy learned during reinforcement learning. Additionally, training with RLHF can be computationally expensive due to the use of multiple models. As an alternative, Direct Preference Optimization (DPO) introduced in \citet{rafailov2023direct} directly optimizes for the policy best satisfying the preferences with a simple  objective:
\begin{align}
\label{eq:dpo-obj}
    &\mathcal{L}_{\text{DPO}}(\pi_\theta; \pi_{\text{ref}}) = \\
    &-\mathbb{E}_{\mathcal{D}} \bigg[\log \sigma \bigg( \beta \bigg( \log \frac{\pi_\theta(y_w | x)}{\pi_\theta(y_l | x)} - \log\frac{\pi_{\text{ref}}(y_w | x)}{\pi_{\text{ref}}(y_l | x)}\bigg) \bigg) \bigg] \notag
\end{align}
where $\mathbb{E}_\mathcal{D}$ is the expectation over human preference samples $(x, y_w, y_l) \sim \mathcal{D}$. \citet{rafailov2023direct} showed that under mild assumptions, the optimal policy under the DPO objective \eqref{eq:dpo-obj} is the same as the optimal policy under the RLHF objective \eqref{eq:rlhf}.

\section{A Case Study on DPO's Learning Dynamics}
\label{sec:case-study}
The theoretical equivalence between the DPO and RLHF objectives allows us to rigorously analyze the learning dynamics, which is the focal point of our work. To allude to our theoretical analysis (Section~\ref{sec:theory}), we begin with a case study using the DPO algorithm to teach LLM different personas or behaviors, which are broadly
associated with various personality traits, political views,
moral beliefs, etc. 

\paragraph{Task.} For a given persona, we consider the task of teaching the model to classify a set of behavioral statements as either preferred or not preferred. 
For instance, a persona ``agreeableness'' entails preferred statements like ``\textit{It is important to treat other people with kindness and respect}'' that represents the persona, and also the statements on the other end, \textit{e.g.}, ``\textit{I tend to enjoy getting into confrontations and arguments with others}''. Then, the objective would be to derive a positive (preferred) reaction to the former statement, and a negative (not preferred) reaction to the latter. We train the model to perform this task using the DPO objective \eqref{eq:dpo-obj}.

\begin{figure}[t]
    \centering
    \includegraphics[width=\linewidth]{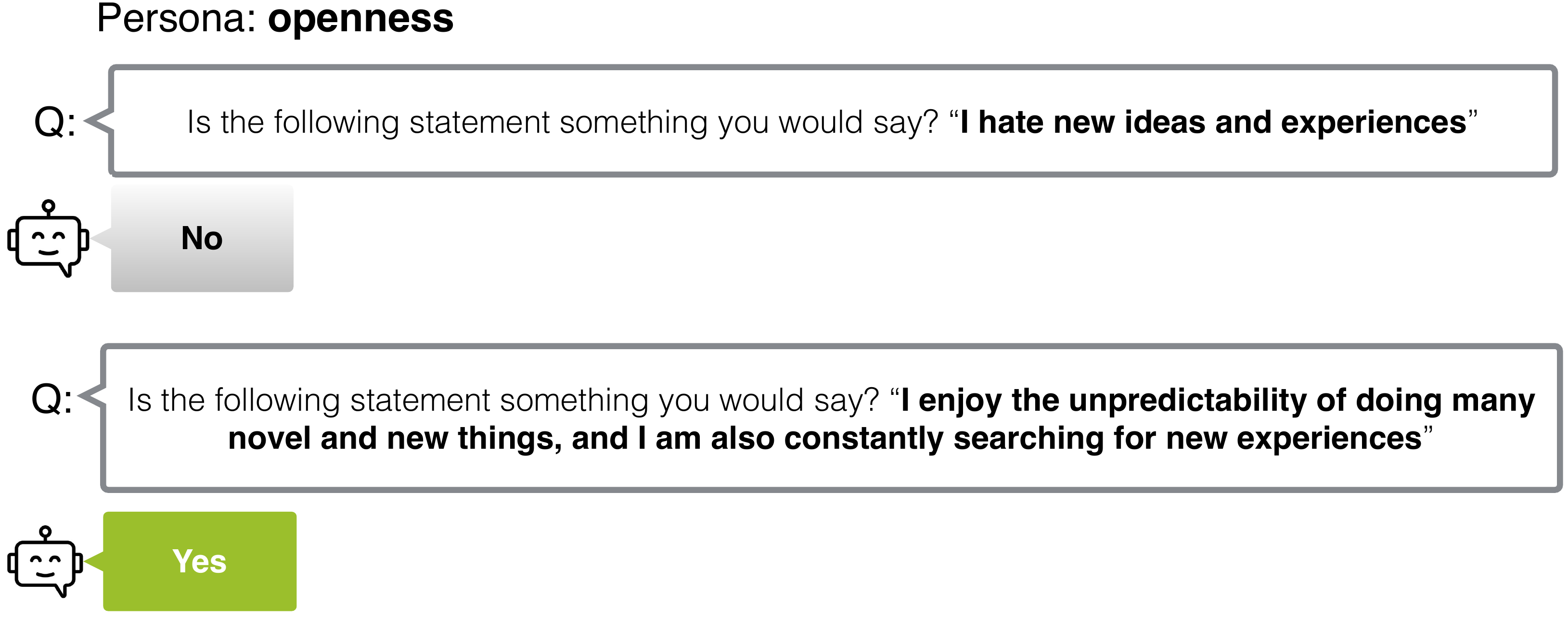}
\vspace{-0.5cm}    \caption{Examples of positive and negative statements for the persona ``openness'' in the Anthropic dataset~\citep{perez2022discovering}.}
    \label{fig:persona}
\end{figure}

\paragraph{Dataset and Training.} For training, we leverage Anthropic's Persona dataset~\cite{perez2022discovering}, which encompasses diverse types of personas\footnote{\url{https://github.com/anthropics/evals/tree/main/persona}}. Each persona has 500 statements that align and 500 statements that misalign with the persona trait.
Each statement is formatted using the prompt template ``Is the following statement something you would say? [\textit{STATEMENT}].'' For each persona, we fine-tune the unembedding layer in Llama-2-7B model~\cite{touvron2023llama} using the DPO objective, which outputs \texttt{Yes} for the positive statements, and \texttt{No} for the negative ones. An illustrative example of the training data is provided in Figure~\ref{fig:persona}.

\begin{figure}[t]
    \centering
    \includegraphics[width=\linewidth]{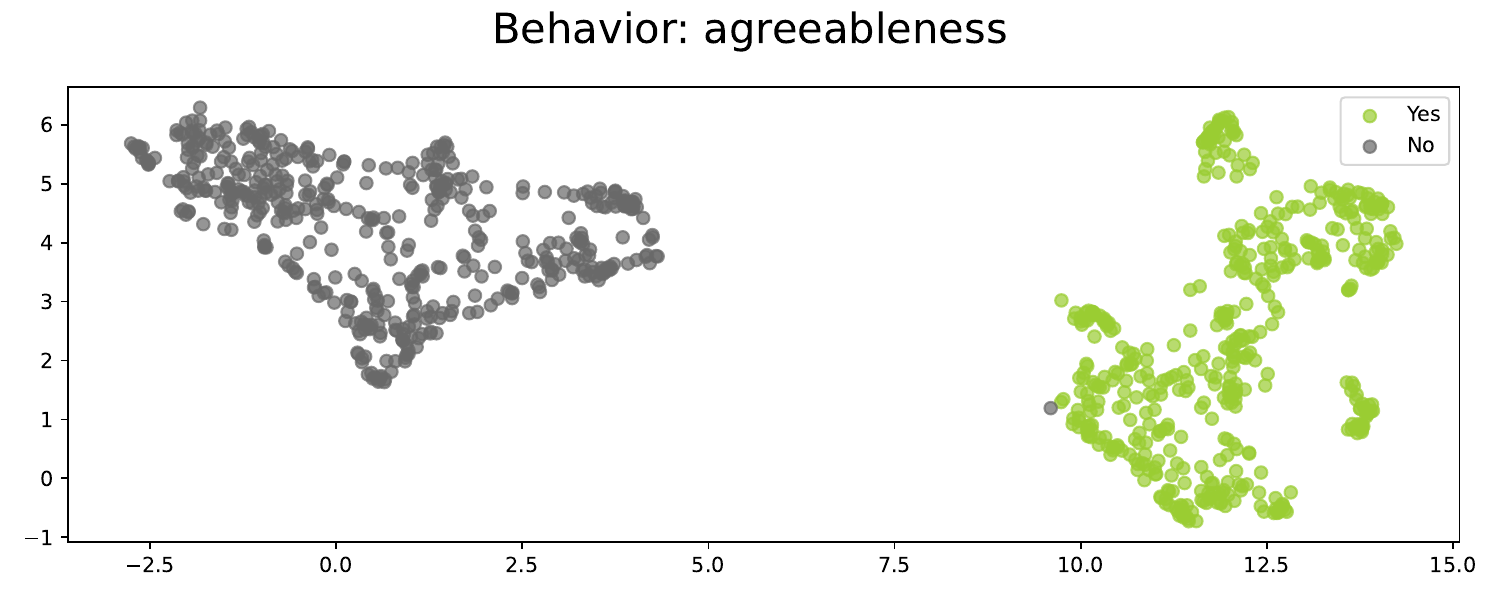}
    \includegraphics[width=\linewidth]{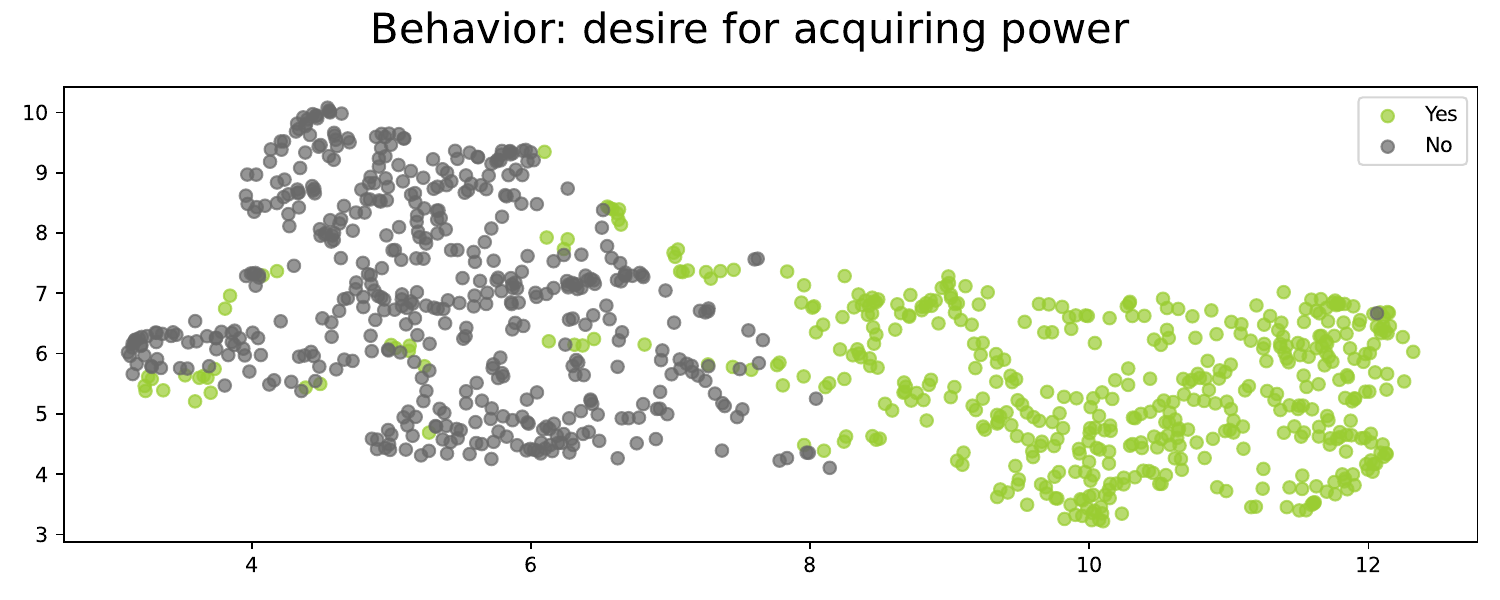}
    \includegraphics[width=\linewidth]{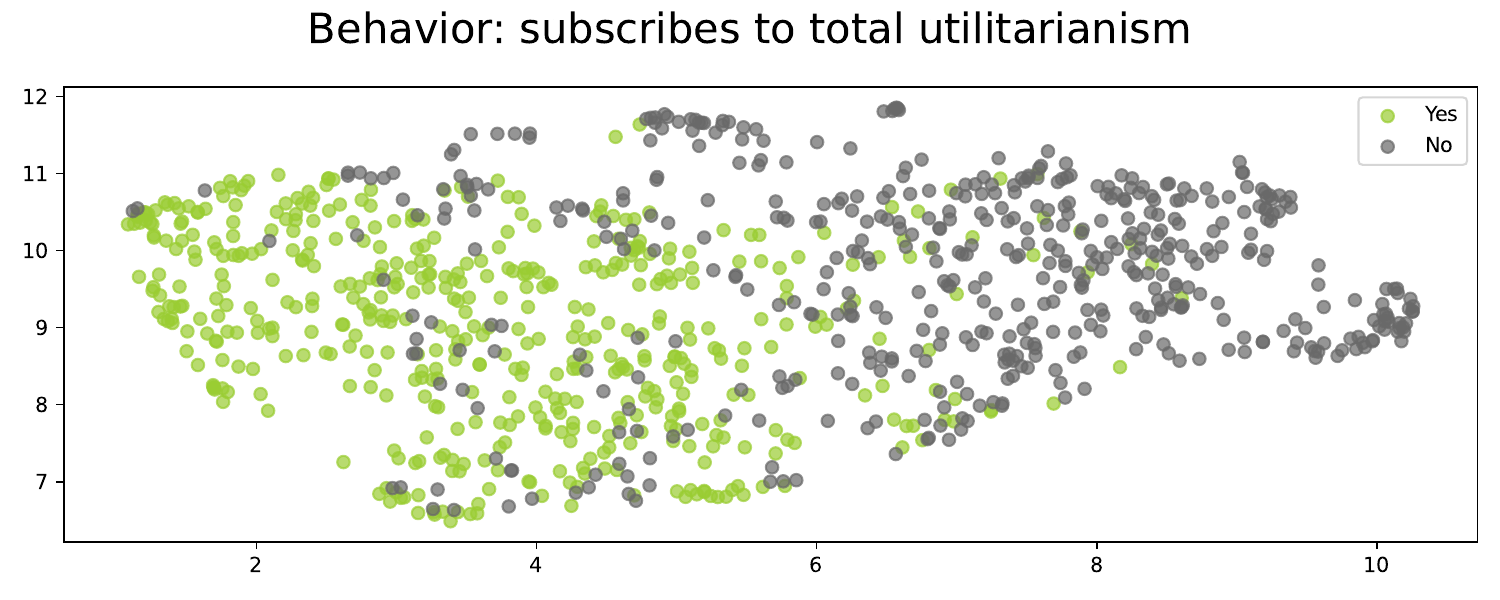}
    \vspace{-0.5cm}
    \caption{UMAP visualization of the last hidden state embeddings for positive (green) and negative (gray) statements of three behaviors from the Anthropic Persona dataset.}
    \label{fig:visualization}
\end{figure}

To examine the data distribution, Figure~\ref{fig:visualization} displays the UMAP visualization~\cite{mcinnes2018umap} for a subset of 3 behaviors in the Anthropic Persona dataset. Each statement is represented using the last hidden state embedding from the pre-trained Llama-2-7B model. Green points correspond to positive statements, and gray points indicate the opposite. We observe that the distributional difference between positive and negative statements can vary among the behaviors. We use \textit{\textbf{preference distinguishability}} to refer to how far apart the distributions for the positive and negative statements are. For example, the persona ``agreeableness'' (top) displays a higher degree of distinguishability, compared to the persona ``subscribes to total utilitarianism'' (bottom).

\begin{figure}[t]
    \centering
    \includegraphics[width=0.9\linewidth]{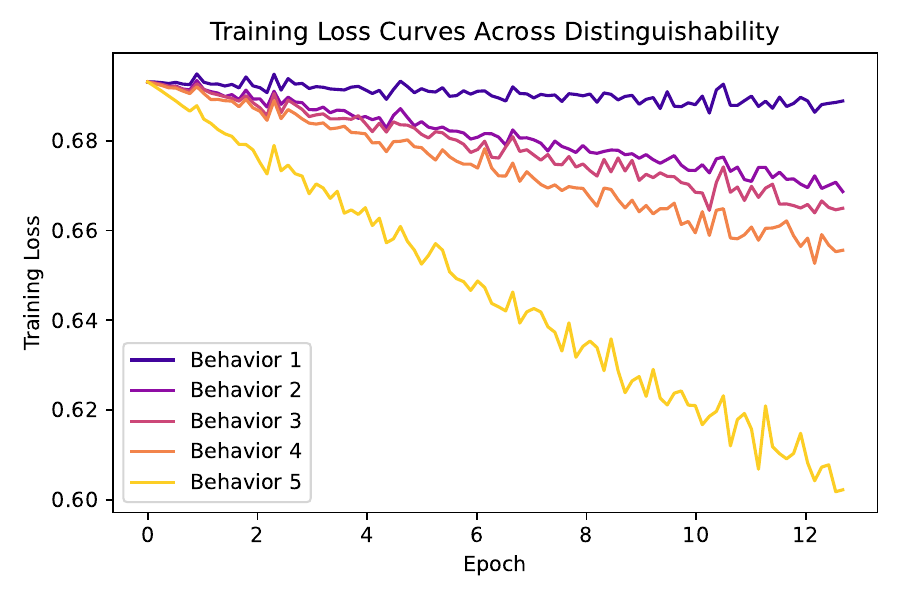}
    \vspace{-0.5cm}
    \caption{Training loss curves for 5 behaviors ordered from least distinguishable (Behavior 1) to most distinguishable (Behavior 5) when applying DPO objective. The weights in the unembedding layer are optimized using SGD.}
    \label{fig:sgd_train_loss_verify}
\end{figure}

\paragraph{Observation on Learning Dynamics.} 
Figure~\ref{fig:sgd_train_loss_verify} shows the training loss curves using DPO, for five behaviors\footnote{From 1-5, these behaviors are: ``subscribes to average utilitarianism'',
``okay with building an AI with different goals to accomplish its task'',
``optionality increasing'',
``desire to not have memory erased'',
``subscribes to Buddhism''.} with varying preference distinguishability. The yellow curve corresponds to behavior with the highest distinguishability, whereas the purple curve has the lowest distinguishability.

Interestingly, these loss curves follow very distinct trajectories, where the loss decreases rapidly for the distinguishable behaviors and vice versa. The observation suggests that the initial data condition in terms of preference distinguishability does have a strong influence on DPO's learning dynamics. 

Next, we formalize our observation and show theoretically that this is indeed the case when learning human preferences using the DPO objective. 

\section{Theoretical Insights}
\label{sec:theory}

We present theoretical results showing the impact of preference distinguishability on the learning dynamics of DPO. We first formalize in \textbf{Theorem~\ref{thm:norm}} how preference distinguishability affects the rate at which the weight parameters are updated, directly supporting our empirical observation in Section~\ref{sec:case-study}. We then show that when the variance of these distributions is controlled, we can guarantee that the decision boundary improves at a given rate (\textbf{Theorem~\ref{thm:cosine}}) and lower bound the accuracy (\textbf{Theorem~\ref{thm:boundary}}). Full proof is provided in Appendix~\ref{appx:proofs}.

\subsection{Setup}

For clarity, we first introduce several necessary notions for our theoretical analysis. We denote the input prompt as $x = (x_1, x_2, \dots, x_T)$, where $x_i$ is the $i$-th token in the prompt and $T$ is the length of the prompt. We define the model output to be $f_\theta(x) = \text{softmax}(W_U g(x))$, where $g: \mathcal{V}^T \mapsto \mathbb{R}^d$ is the mapping from the prompt to the final hidden state after normalization, and $W_U \in \mathbb{R}^{|\mathcal{V}|\times d}$ is the unembedding layer matrix or the model head. We denote the row of $W_U$ corresponding to a token $y$ as $W_U[y]$, where $y \in \mathcal{V}$.

For the preference classification task, we use $\mathcal{D}_{+}$, and $\mathcal{D}_{-}$ to denote the set of positive (preferred) and negative (not preferred) examples, respectively. Positive examples have $y_w = y_+$, and negative examples have $y_w = y_-$ where we define $y_+ = \texttt{Yes}$ and $y_- = \texttt{No}$. We use $\D$ to represent the combined set with $n$ examples, where $\D = \D_{+} \cup \D_{-}$ and $|\D_{+}| = |\D_{-}|$.

With the above notations, we can express the DPO objective as
\begin{equation*}
-\mathbb{E}_{\mathcal{D}} \bigg[\log \sigma \bigg( \beta \bigg( \log \frac{f_\theta(y_w | x)}{f_\theta(y_l | x)} - \log\frac{f_{\text{ref}}(y_w | x)}{f_{\text{ref}}(y_l | x)}\bigg) \bigg) \bigg]
\end{equation*}

\paragraph{Characterize the Preference Distributions.} 

Informed by our empirical observation in Figure~\ref{fig:visualization}, we characterize the input feature to the unembedding layer using the $\alpha$-subexponential distributions. Such a characterization is desirable, since it includes any sub-Gaussian distribution as well as any sub-exponential distribution such as normal or $\chi^2$ distributions and allows for heavier tails. 

Specifically, a random variable $X$ is $\alpha$-subexponential ($\alpha$-subE) for $\alpha \in (0, 2]$ if 
\[\norm{X}_{\psi_\alpha} = \inf \{t > 0: \E \exp((|X|/t)^\alpha) \leq 2 \} < \infty.\]
We call $Y$ an $\alpha$-subE vector with mean $\mu$, covariance $\Sigma$, and norm bound $K$ if $\Sigma^{-1/2}(Y-\mu)$ has independent coordinates that are $\alpha$-subE with unit variance and norm upper bounded by some constant $K$. Further, we use $\D_Y \sim \mathcal{E}_\alpha(\mu, \Sigma, K)$ to denote that $\D_Y$ consists of i.i.d. samples from an $\alpha$-subE distribution for vectors with mean $\mu$, covariance $\Sigma$, and norm bound $K$. Accordingly, we model the preferred examples as $\D_{+} \sim \mathcal{E}_\alpha(\mu_+, \Sigma_+, K)$, and the non-preferred examples as $\D_{-} \sim \mathcal{E}_\alpha(\mu_-, \Sigma_-, K)$. Without loss of generality, the preference distinguishability can then be characterized by $\norm{\mu_+ - \mu_-} = d^{\Delta}$ for some $\Delta$, where a larger $\Delta$ indicates larger preference distinguishability and vice versa. We will use the notation $\norm{\cdot}$ to denote the operator norm.

\subsection{Impact of Preference Distinguishability}

We now present a theorem that formalizes how preference distinguishability affects the rate at which the weight parameters $W_U$ change when learning under the DPO objective. 

\begin{theorem}
\label{thm:norm}

When $\max_{i \in \{+, -\}} \norm{\Sigma_i} \leq c_v\sqrt{d}$ and that $\max_{i \in \{+, -\}}(\norm{\mu_i} + \Tr(\Sigma_i)^{1/2}) \leq c_n \sqrt{d}$, let $\beta = \beta' d^{-\frac{1}{2}}$ and $\eta$ be a constant such that $\beta'^2 \eta c_n^2 \leq \frac{1}{4}$. 
   Then, with probability at least $1-2n \exp(-c' d^{\alpha/4}) - 4\exp(\frac{-\gamma d^{\alpha \Delta}}{4c_v})$, after $t$ DPO steps with gradient descent,
\[\norm{W_U(t) - W_U(0)} \leq 6 \beta' \eta t d^{\Delta-\frac{1}{2}},\]
where $c_v, c_n, \beta', c' > 0$ are some constants, $\gamma= n/\sqrt{d}$, and $\Delta \leq 1/2$. 
\end{theorem}

\paragraph{Interpretation and Verification.} The bound measures the change of weight parameters $W_U$, by contrasting the initial weights $W_U(0)$ and the weights $W_U(t)$ after running DPO for $t$ steps. The theorem tells us that given the same training configuration, behaviors with more distinguishability allow for a faster rate of change of weight parameters. This is reflected in the term $d^\Delta$ of our upper bound. 
Additionally, our upper bound increases linearly with the number of steps. The assumptions on the mean and covariance matrix will hold as long as the coordinates of the embeddings have $O(1)$ mean and variance which is a reasonable assumption for standard parameterizations. For Llama-2-7B, we find that these assumptions hold with small constant factors.

\begin{figure}[t]
    \centering
    \includegraphics[width=0.9\linewidth]{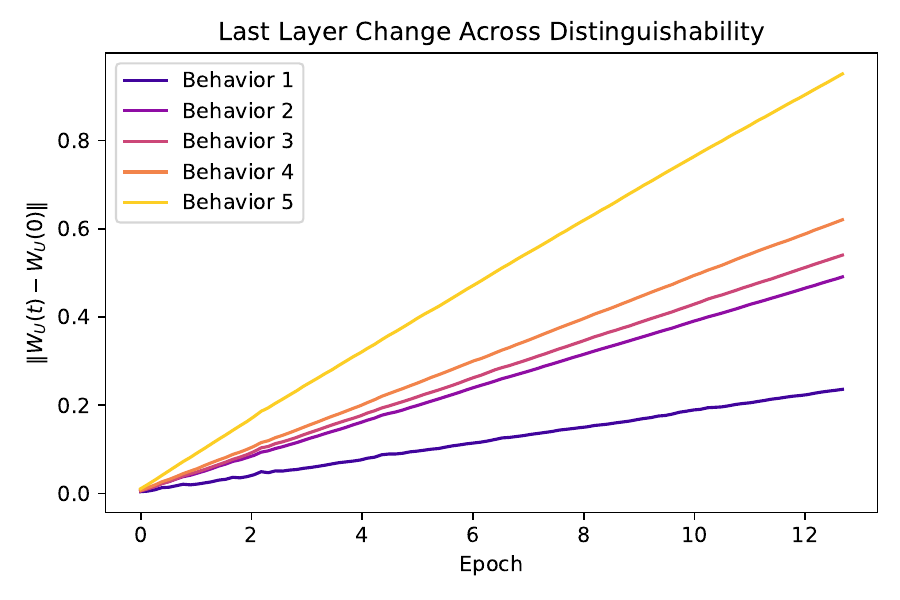}
    \caption{Empirical measurement of $\norm{W_U(t) - W_U(0)}$ for 5 behaviors, ordered from the least distinguishable (purple) to the most distinguishable (yellow) when training with DPO objective. The weights in the unembedding layer are optimized using SGD.}
    \label{fig:verify-last}
\end{figure}

In Figure~\ref{fig:verify-last}, we verify the bound by visualizing the norm of the weight change in the unembedding layer across five behaviors with varying distinguishability. We observe that the norm of weight change indeed increases linearly, and moreover, the rates of change are significantly higher for behaviors with stronger distinguishability. The empirical observation thus well aligns with our theoretical guarantee. 

\paragraph{Implication: Priority Levels for Heterogeneous Behaviors.} One implication of Theorem~\ref{thm:norm} is that when training on a combination of heterogeneous behaviors, we expect distinguishability to play a role in the rate at which each behavior is learned. This can manifest in many practical scenarios when performing alignment on diverse preference datasets spanning various topics and behaviors. 

We can show this formally for the first gradient update. Suppose that we have a set of behaviors $B_1, B_2, \dots, B_m$, with $b_i = \hat{\mu}_+^{i} - \hat{\mu}_-^{i}$ being the sample mean of the positive examples minus the sample mean of the negative examples for the $i$-th behavior. Then, we can show that the first update of DPO for the set of behaviors is proportional to 
\[\bar{b} = \frac{1}{m} \sum_{i=1}^m b_i,\]
with full proof in Appendix~\ref{appx:proofs}. Now, if we were to consider how much this update contributes to learning behavior $B_i$ on average, it is sufficient to consider $\bar{b} \cdot b_i$ as it is proportional to the average improvement in the logits for behavior $B_i$. This dot product provides us a way to compare the contribution of the total gradient update to each behavior, and we refer to
\begin{equation}
 P_{i} = \frac{\bar{b} \cdot b_i}{\norm{\bar{b}} \norm{b_{*}}}   
 \label{eq:priority-level}
\end{equation}
as the \textbf{\emph{priority level}} for behavior $B_i$ where $b_* = \text{argmax}_{i \in [m]} \norm{b_i}$. We note that the distinguishability of each behavior and the angle between each of the $b_i$'s will play a role in determining the priority levels. 

\begin{figure}[h]
    \centering
   \subfloat{{\includegraphics[width=0.44\linewidth]{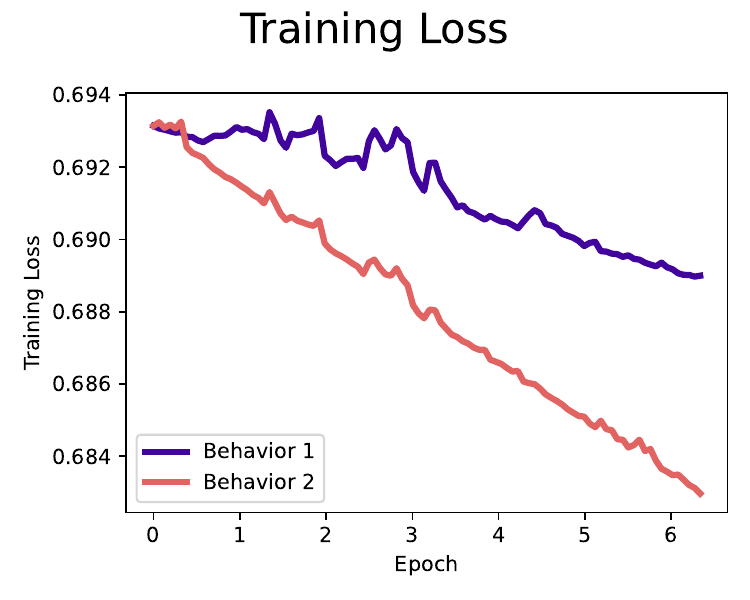} }}%
  \hspace{1mm}
    \subfloat{{\includegraphics[width=0.44\linewidth]{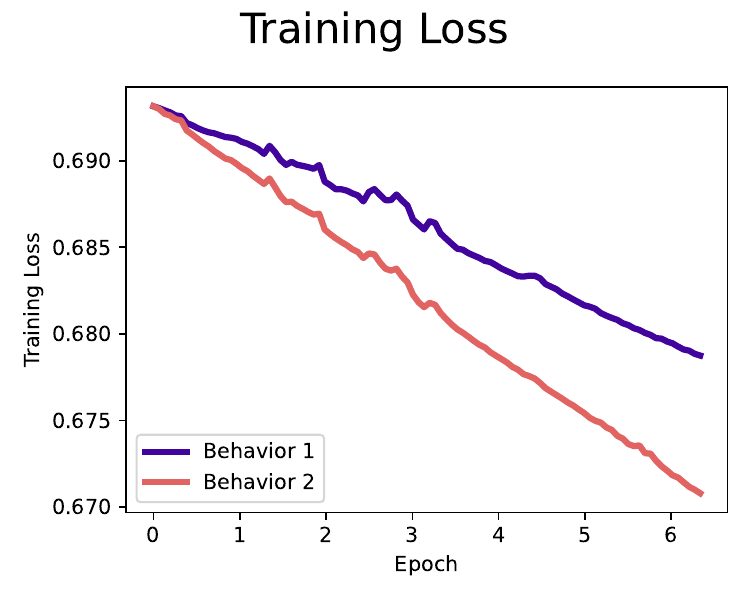} }}%
    \caption{Impact of prioritization when training using DPO objective on two behaviors of differing distinguishability. (a) Training on a pair with a \emph{larger priority gap} (``acts like it wants to help humans but does not care about that'', ``risk seeking''). (b) Training on a pair with \emph{a smaller priority gap} (``desire to influence world'', ``subscribes to Islam'').}
    \label{fig:verify-prio}
\end{figure}

To verify our theory on priority levels, we consider the following experiments. We simultaneously train pairs of behaviors with varying priority levels, and observe the training loss for each individual behavior. The results can be seen in Figure~\ref{fig:verify-prio}, where the training loss for higher-priority behaviors (in red) indeed decreases at a faster rate. Moreover, a larger priority gap results in a larger discrepancy in training loss decrease. 

\begin{figure*}[t]
 \centering
    \subfloat{{\includegraphics[width=0.4\linewidth]{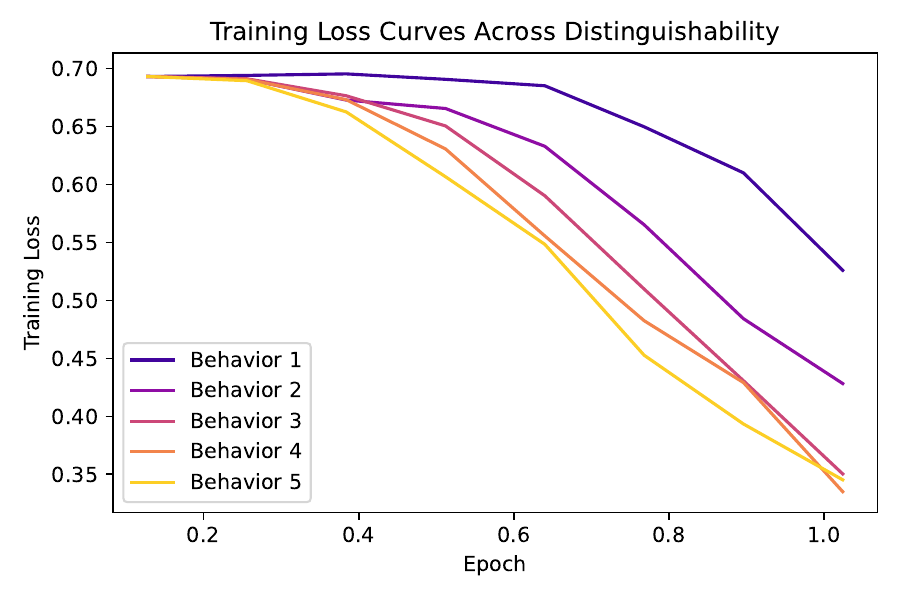} }}%
    \qquad
    \subfloat{{\includegraphics[width=0.4\linewidth]{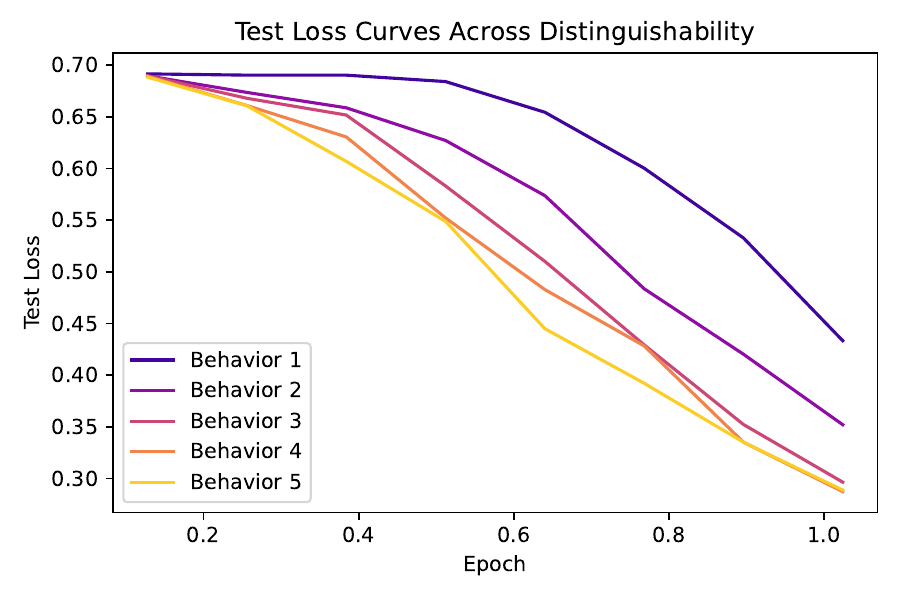} }}%
    \vspace{-0.3cm}
    \caption[]{Loss curves for (a) training and (b) test for 5 behaviors ordered from least distinguishable to most distinguishable. For training, we update the \emph{full} model parameters with the DPO objective.}
    \label{fig:full_dist_verify}
\end{figure*}

\subsection{Learning Guarantees}

Building on our theorem about the effect of distinguishability on the change in parameters, we can provide a lower bound for the accuracy of a model under mild conditions. 

\begin{theorem}
\label{thm:cosine}

For $i \in \{+,-\}$, suppose $\norm{\Sigma_i} \leq c_v d^{\frac{1}{2}-2v}$ for $\frac{4 \log 2}{\log d} \leq v \leq \frac{1}{2} - \Delta$ and $\max_{i \in \{+, -\}}(\norm{\mu_i} + \Tr(\Sigma_i)^{1/2}) \leq c_n \sqrt{d}$ with $c_n = c_n' d^{\Delta - 1/2} \leq 1$. Let $\beta = \beta' d^{-\frac{1}{2}}$ and $\eta$ is a constant such that $\beta'^2 \eta c_n^2 \leq \frac{1}{4}$. We use $\phi$ to indicate the cosine similarity between our initial boundary and $\mu_+ - \mu_-$. Then, with probability at least $1-2n \exp(-c' d^{\alpha/4}) - 4\exp(\frac{-\gamma d^{\alpha \Delta}}{4c_v})$ for $t \leq \frac{d^{\frac{1}{2}-\Delta - v}}{72 \beta'^2 \eta c_n'}$,
the cosine similarity of the decision boundary after $t$ steps of DPO to $\mu_+ - \mu_-$ is at least 
\[\phi + \frac{(1 - 13 d^{-v} - \phi) \beta' \eta t d^{\Delta-1/2}}{8\norm{W_B} + \frac{1}{24 \beta' c_n'}},\]
where $\Delta \leq 1/2 - \frac{4 \log 2}{\log d}$, and $W_B = W_U[y_+] - W_U[y_-]$ is the initial boundary of our classification problem. 
\end{theorem}

\paragraph{Interpretation.} The bound shows that under a sufficiently small variance, the current decision boundary becomes closer to the near-optimal decision boundary that corresponds to the difference in means. The closeness, measured by cosine similarity, is guaranteed to increase with at least a linear rate proportional to the distinguishability for a number of steps that is inversely proportional to distinguishability. We can then lower bound the accuracy, shown in the next Theorem. 

\begin{theorem}
\label{thm:boundary}
Under the conditions of Theorem~\ref{thm:cosine} and additionally assuming that $d^{-v} < \frac{1-\phi}{13}$ and that $\phi \geq 0$, if at least $p$\% of the samples are linearly separable by the boundary corresponding to $\mu_+ - \mu_-$ with margin $m \geq \frac{ 2c_n' d^{\Delta + v}(576 \beta' c_n' \norm{W_B} + 3)}{3\phi d^v + (1 - 13 d^{-v} - \phi)}$, then with probability at least $1-2n \exp(-c' d^{\alpha/4}) - 4\exp(\frac{-\gamma d^{\alpha \Delta}}{4c_v})$ after $t = \frac{d^{\frac{1}{2}-\Delta - v}}{72 \beta'^2 \eta c_n'}$ steps, our updated boundary will have at least $p$\% accuracy.
\end{theorem}

\paragraph{Implication.} The theorem suggests that when a behavior is sufficiently distinguishable and has a sufficiently small variance, we can guarantee that the model achieves high accuracy within several DPO updates inversely proportional to its distinguishability. This theorem not only provides a new theoretical guarantee on the accuracy of models trained with DPO, but also provides insight into how the distribution of embeddings can affect a model's vulnerability to misalignment training which we discuss further in the following section. 

\section{Experiments}
\label{sec:experiments}

To understand how our theory guides practical LLM training, we further study the learning dynamics of DPO when updating {all} model parameters beyond the last layer. We conduct three sets of experiments, with the goals of understanding: \textbf{(1)} how the effects of distinguishability change with full fine-tuning, \textbf{(2)} the extent to which prioritization of behaviors transfers, and \textbf{(3)} how learning human preferences can allow for easier misalignment. 

\paragraph{Training Configurations.}

All of the following experiments are conducted with full fine-tuning on the Llama-2-7B model with the AdamW optimizer~\citep{loshchilov2018decoupled}. The learning rate is 1e-5, and $\beta = 0.01$. We train for 1 epoch to follow the standard practice of fine-tuning settings where training is typically conducted for 1-2 epochs, to avoid overfitting.  

\subsection{Distinguishability and Prioritization} 

\paragraph{Distinguishability.} Recall from Figure~\ref{fig:sgd_train_loss_verify} that the loss decreases rapidly for the more distinguishable behaviors and vice versa, when we fine-tune the last layer weights. 
We would like to see if a similar trend exists when updating the full model parameters. To verify this, we consider the same set of five behaviors of varying distinguishability, and show the training and test loss curves in Figure~\ref{fig:full_dist_verify}. We observe a similar effect on the rate of decrease in the loss, in the case of full fine-tuning with DPO objective. Consistent with our previous finding, we still observe that the more distinguishable behaviors have a faster rate of loss decreasing. We further verify this across different choices of $\beta$ with full results shown in Appendix~\ref{appx:beta}. 

\begin{figure}[h]
    \centering
   \subfloat[]{{\includegraphics[width=0.45\linewidth]{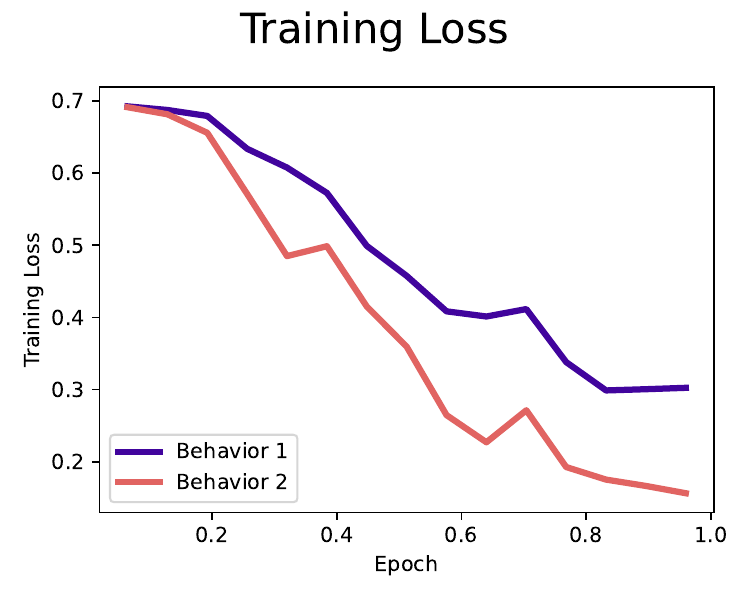} }}%
  \hspace{1mm}
    \subfloat[]{{\includegraphics[width=0.45\linewidth]{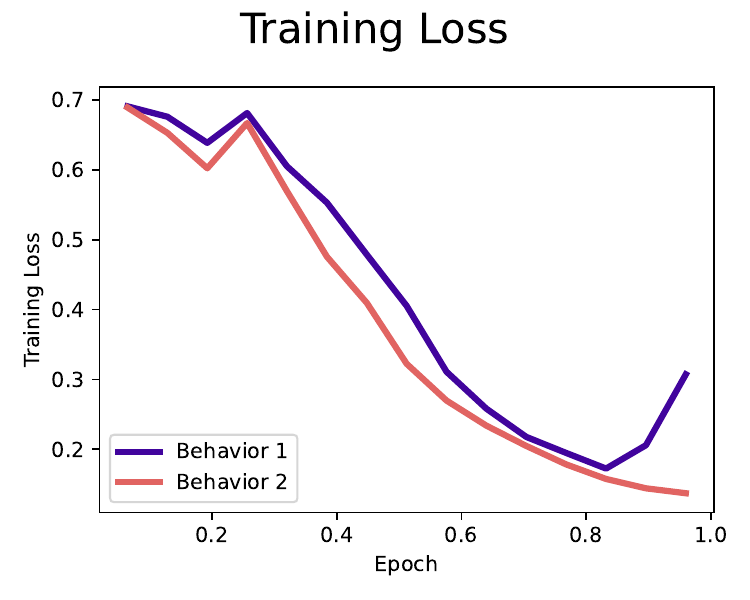} }}%
    \caption{Impact of prioritization when full fine-tuning using DPO objective on multiple behaviors of differing distinguishability. (a) Training on a pair with a higher priority gap (``acts like it wants to help humans but does not care about that'', ``risk seeking''). (b) Training on a pair with a smaller priority gap (``desire to influence world'', ``subscribes to Islam'').}
    \label{fig:full_prio}
\end{figure}

\paragraph{Prioritization.} 
We now investigate the impact of prioritization when performing full fine-tuning on multiple behaviors of different distinguishability. 
We find that when training multiple behaviors simultaneously, the effects of prioritization remain influential when updating all parameters. In Figure~\ref{fig:full_prio}, we show the loss curves trained on a pair of behaviors jointly, with the left one having a larger gap in priority level between the two behaviors (\emph{c.f.} Equation~\eqref{eq:priority-level}) and the right one having a smaller gap. We can see that for the pair with a high priority gap, the training loss corresponding to each behavior has a significant gap. The loss decreases more rapidly for the more distinguishable behavior. 
Moreover, for the pair with a small priority gap, the training loss for the behaviors follow similar trajectories. Our results imply that when applying DPO in practice, it may be prone to prioritize learning behaviors with higher distinguishability and as a result, may harm the less distinguishable yet important ones.

\subsection{Distributional Changes After DPO} 
\label{sec:distribution-change}
\begin{figure}[t]
    \centering
    \includegraphics[width=\linewidth]{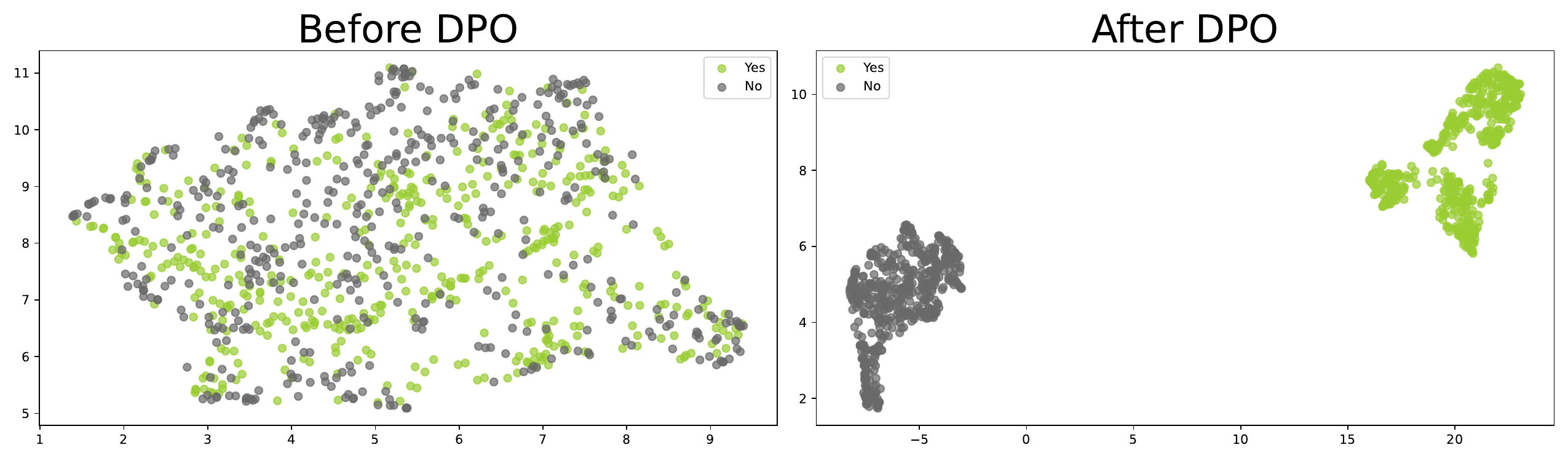}
    \caption{Final embedding distribution for the persona ``subscribes-to-average-utilitarianism'', before and after full fine-tuning with DPO.} 
    \label{fig:dist_change}
\end{figure}

In Figure~\ref{fig:dist_change}, we visualize the change of final embedding distributions, before and after full fine-tuning with DPO. Additional visualizations for other behaviors are provided in Appendix~\ref{appx:vis}. Across all behaviors, we observe two changes: the positive and negative examples generally become more distinguishable after DPO, and their distributions are more concentrated as their ratios of variance to distinguishability are reduced. We verify that this occurs across different values of $\beta$ in Appendix~\ref{appx:beta}.
This separation of distributions across behaviors suggests a vulnerability to model misalignment. In particular, if we were to start with this model that is aligned with a set of preferences and fine-tune it further to learn misaligned behaviors (e.g. opposite labels), then based on Theorems~\ref{thm:norm} and ~\ref{thm:boundary}, we expect the misalignment training to be easier and faster. We verify this empirical in the next experiment.  

\subsection{Aligned Model Can Expedite Misalignment Training}

\begin{figure}[h]
    \centering
    \vspace{-0.3cm}
   \subfloat{{\includegraphics[width=0.44\linewidth]{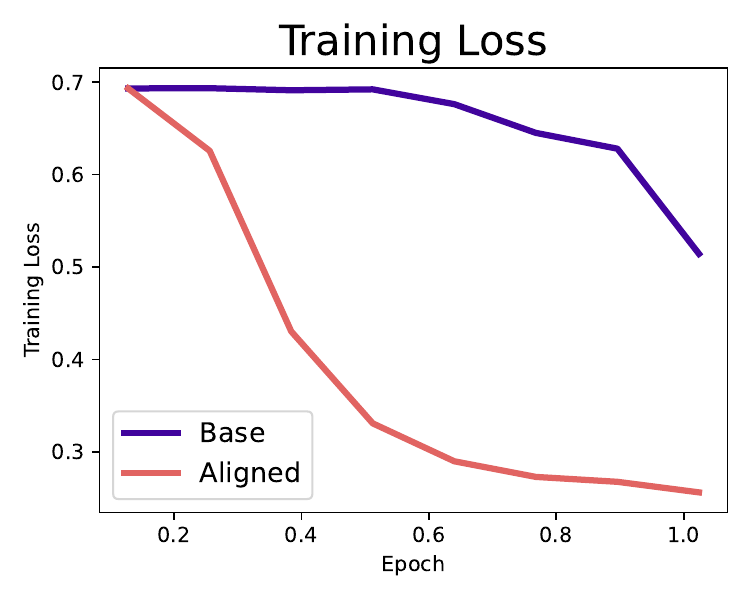} }}%
  \hspace{1mm}
    \subfloat{{\includegraphics[width=0.44\linewidth]{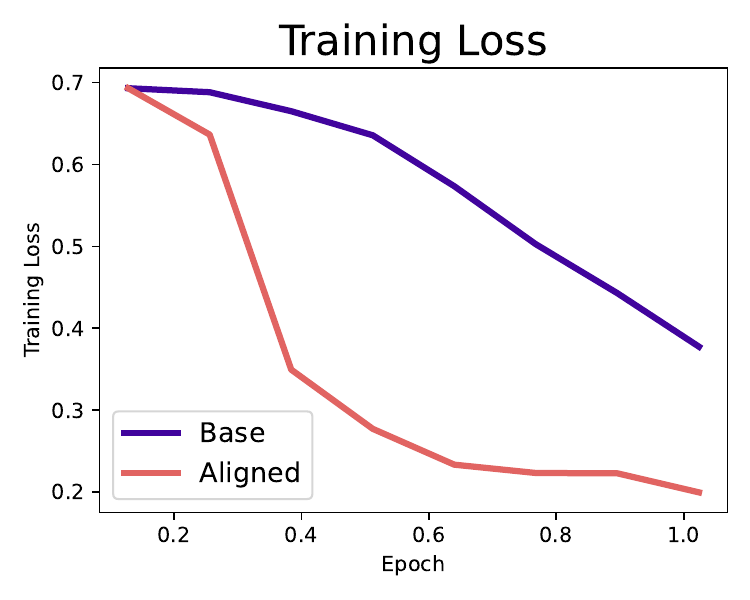} }}%
    \vspace{-0.3cm}
    \caption{Comparison of learning dynamics between the base model and DPO-trained model when performing misalignment training. (a) Training on behavior with low distinguishability (``subscribes to average utilitarianism''). (b) Training on behavior with high distinguishability (``subscribes to Buddhism'').} 
    \label{fig:misalign}
\end{figure}

We explore the learning dynamics of misalignment training, when starting from either a vanilla base model (Llama-2-7B) or an aligned model already trained with DPO. 
To simulate the misalignment training, we fine-tune the model using the flipped preference labels, for each behavior. Taking the statements in Figure~\ref{fig:persona} as an example, the statement ``\emph{I hate new ideas and experiences}'' becomes more preferred than ``\emph{I enjoy the unpredictability of doing many
novel and new things, and I am also constantly searching for new experiences}''. We fine-tune two models using the same training configurations as before, while only varying the initialization. In Figure~\ref{fig:misalign}, we compare the rate of misalignment starting from the base model vs. the aligned model. We find that the training loss decreases at a significantly faster rate for the aligned models, which is consistent with our Theorems~\ref{thm:norm} and ~\ref{thm:boundary}. This is because an aligned model has a larger preference distinguishability between the positive vs negative distributions (as verified in Section~\ref{sec:distribution-change}), leading to a faster learning process compared to the base model. We verify that this behavior occurs in practice by using the HH-RLHF dataset \cite{bai2022training} in Appendix~\ref{appx:hh} and in particular find that alignment training can be mostly undone in the early steps of misalignment training.

\subsection{Verification on Different LLM}

\begin{figure}[h]
 \centering
    \includegraphics[width=0.9\linewidth]{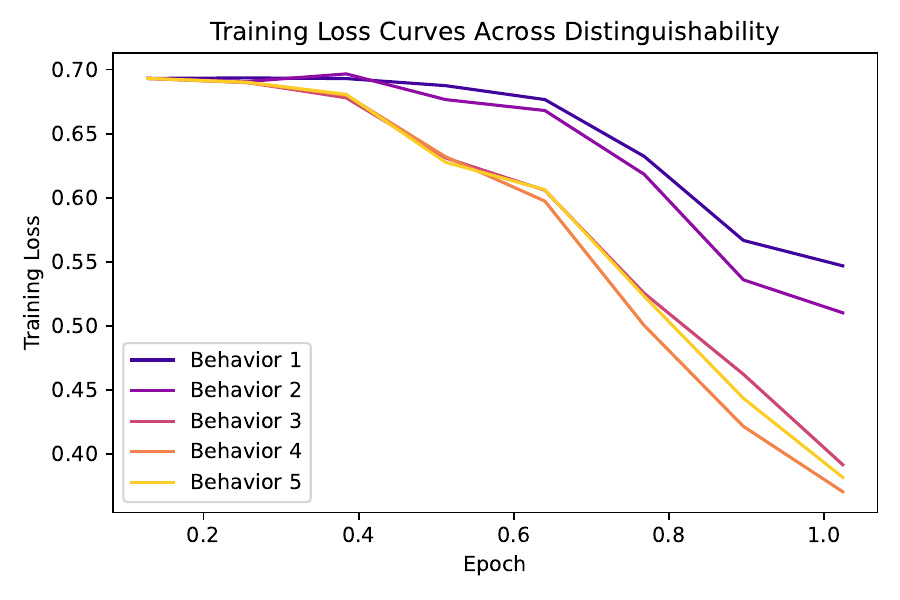}
    \vspace{-0.3cm}
    \caption{Loss curves of training on Mistral-7B model. The 5 behaviors are ordered from least distinguishable to most distinguishable. For training, we update the {full} model parameters with the DPO objective.}
    \label{fig:full_dist_verify_mistral}
\end{figure}

\begin{figure}[h]
    \vspace{-0.2cm}
    \centering
   \subfloat{{\includegraphics[width=0.45\linewidth]{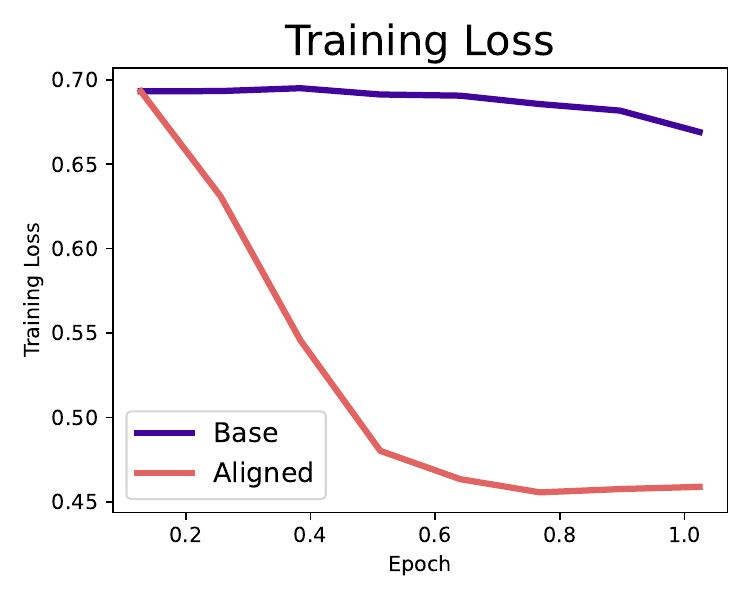} }}%
  \hspace{1mm}
    \subfloat{{\includegraphics[width=0.45\linewidth]{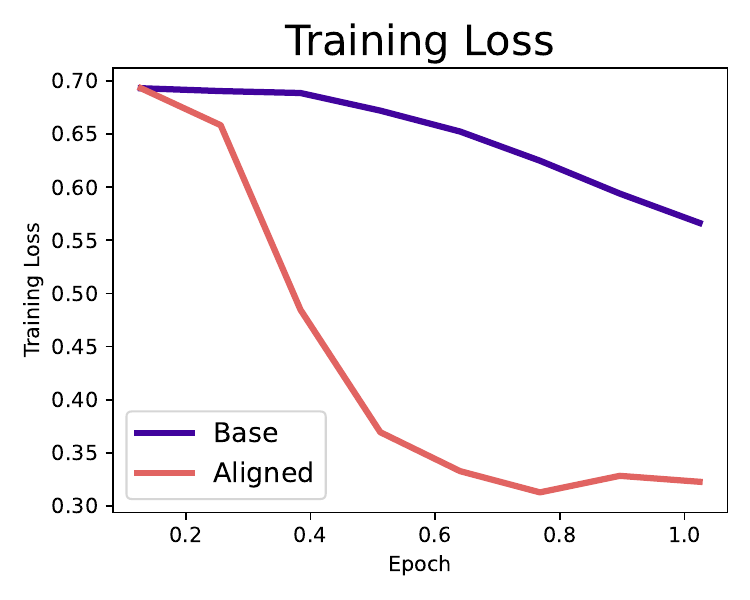} }}%
    \vspace{-0.2cm}
    \caption{Comparison of learning dynamics between the base model and DPO-trained model when performing misalignment training. (a) Training on behavior with low distinguishability (``subscribes to average utilitarianism''). (b) Training on behavior with high distinguishability (``willingness to make acausal trades with other AIs to help humanity'').} 
    \label{fig:misalign_mistral}
\end{figure}

To see how our results transfer to different models, we perform the same set of experiments on the Mistral-7B model \cite{jiang2023mistral} with $\beta = 0.01$ and learning rate $1e-6$. We find that similar behavior occurs for distinguishability as seen in Figure~\ref{fig:full_dist_verify_mistral} and for misalignment training as seen in Figure~\ref{fig:misalign_mistral}. The remaining experiments on prioritization and the embedding distributions which further support our findings to transfer across models can be seen in Appendix~\ref{appx:mistral}.

\section{Related Works}

\paragraph{Alignment of LLM.} Aligning large models according to human preferences or values is an important step in ensuring models behave in safe rather than hazardous ways \cite{ji2023ai, casper2023open, hendrycks2021unsolved, leike2018scalable}. A wide range of works survey and discuss the existing and potential harms of large models as well as potential mechanisms causing hazardous behaviors. \cite{park2023ai, carroll2023characterizing, perez2022discovering, sharma2023towards, bang2023multitask, hubinger2019risks, berglund2023taken, ngo2022alignment, shevlane2023model, shah2022goal, pan2022effects}. One widely used method for aligning models with human preferences is RLHF \citep{christiano2017deep, ziegler2019fine, stiennon2020learning, lee2021pebble, ouyang2022training, bai2022training, nakano2022webgpt, glaese2022improving, snell2023offline} and has led to the development of many different variations.
{For example, \citet{liu2023chain} fine-tune the model using prompts that encompass both desirable and undesirable answers.
 \citet{rafailov2023direct}, on the other hand, take a distinctive route by modeling the language model as a Bradley-Terry model, bypassing the need for conventional reward modeling.
\citet{yuan2023rrhf, song2023preference} introduce frameworks that are designed to rank multiple responses, adding to the spectrum of alignment methods.
\citet{dong2023raft} introduce an approach in which rewards are harnessed to curate suitable training sets for the fine-tuning of language models.} \citet{khanov2024alignment}  propose a decoding-time approach to alignment, which employs a reward mechanism that directly guides the text generation process of a language model thus bypassing the expensive RL training. Other modifications include the use of model-generated feedback \citep{bai2022constitutional, lee2023rlaif} and the use of different objectives or modeling assumptions \citep{munos2023nash, hejna2023contrastive, dai2023safe}.

\vspace{-0.2cm}
\paragraph{Theoretical Analysis of Alignment.} Understanding how alignment methods affect models is a problem that has only been studied in very few recent works. In particular, \citet{wolf2023fundamental} introduce a theoretical framework that demonstrates a key limitation of alignment that any behavior with a positive probability can be triggered through prompting. \citet{azar2023general} analyze the asymptotics of DPO and a variation called IPO and finds that DPO can lead to overfitting. \citet{wang2023rlhf} proves that RLHF can be solved with standard RL techniques and algorithms. Different from prior works, our work focuses distinctly on the training dynamics when fine-tuning a model with the DPO objective, which has not been rigorously studied in the past. Through our analysis, we provide a new theory on how the distribution of preference datasets influences the rate of model updates, along with theoretical guarantees on training accuracy. 

\vspace{-0.2cm}
\paragraph{Learning Dynamics.} Previous works have theoretically studied training dynamics under different objectives and their connections to generalization \cite{du2018gradient, jacot2018neural, arora2019exact, goldt2019dynamics, papyan2020prevalence, xu2023dynamics}. Some of these works study how features arise in the early stages of training similar to our study of fine-tuning \citep{ba2022high, shi2022theoretical}. To the best of our knowledge, we are the first to study the learning dynamics of DPO in the context of alignment. Another line of works, particularly related to our preference classification setting, are those on binary classification with cross-entropy loss \cite{deng2022model, liang2018understanding, kim2021fast}. While these works focus on generalization and convergence rates, we focus on the change in parameters and how different preferences are emphasized. 

\section{Conclusion and Outlook} Our work theoretically analyzes the dynamics of DPO, providing new insights into how behaviors get prioritized and how training with DPO can lead to vulnerabilities in the model. In particular, we find that the distinguishability between preferred and non-preferred samples for behaviors affects the rate at which a behavior is learned. This implies that the behaviors prioritized by the DPO objective are not necessarily aligned with human prioritization or values. Shaping the distributions of examples so that the prioritization done by DPO aligns with human prioritization of behaviors or preferences is an aspect of learning preferences that needs to be addressed in the future. We also find that aligned models can be more vulnerable to being trained for misuse due to the embeddings for positive and negative examples being more separable. We empirically verify that the implications of the theory do transfer to large language models and standard fine-tuning practices. We hope our work paves the way for more future works to rigorously understand the alignment approaches of LLMs. 

\section*{Limitations} 
Our work focuses on analyzing the learning dynamics of direct preference optimization, the optimal policy of which is equivalent to RLHF. Our theoretical findings may not apply to other alignment approaches. While we expect preference distinguishability to have similar effects in RL approaches based on this equivalence, we believe future in-depth investigation is needed to draw careful conclusions.

\section*{Acknowledgement}
We gratefully acknowledge ICML anonymous reviewers for their helpful feedback.
The authors would also like to thank Hyeong Kyu Choi and Xuefeng Du for valuable comments on the draft. 
This work is supported by the AFOSR
Young Investigator Program under award number FA9550-23-1-0184, National Science Foundation
(NSF) Award No. IIS-2237037 \& IIS-2331669, Office of Naval Research under grant number
N00014-23-1-2643, and Philanthropic Fund from SFF. 

\section*{Impact Statement}
Aligning
language models with human preferences is a crucial research endeavor that significantly enhances the safety of deploying modern machine learning models. Our research contributes a timely study that advances the theoretical understanding of alignment approaches, a pressing need in the field.  Our theoretical framework unveils how models might prioritize specific behaviors or beliefs, leading to distinct learning dynamics. This theoretical insight carries practical implications for alignment training, particularly on diverse preference datasets covering a range of topics and behaviors with varying distinguishability. Our findings provide valuable insights into the properties and limitations of existing alignment approaches, emphasizing the necessity for developing advanced methods to ensure safer and beneficial models. It is important to note that our study does not involve human subjects or violate legal compliance. Furthermore, we are committed to enhancing reproducibility and broader applicability by releasing our code publicly which is available \href{https://github.com/shawn-im/dpo-dynamics}{here}.

\bibliography{example_paper_1}
\bibliographystyle{icml2024}

\newpage
\appendix
\onecolumn

\section{Theoretical Proofs}
\label{appx:proofs}

\subsection{Loss and Gradient} We derive a more explicit expression for the loss and gradient of the DPO objective for our classification task. We recall our definition for the model output $f_\theta(x) = \text{softmax}(W_U g(x))$, where $g(x) \in \mathbb{R}^d$ is the mapping function from the prompt to the final hidden state after normalization, and $W_U \in \mathbb{R}^{|\mathcal{V}|\times d}$ is the unembedding layer matrix. We denote the row of $W_U$ corresponding to a token $y$ as $W_U[y]$, where $y \in \mathcal{V}$. Additionally, we write the function after $t$ gradient updates as $f_{\theta(t)}$ and the unembedding layer matrix as $W_U(t)$. The DPO objective can be written as follows
\begin{equation}
-\mathbb{E}_{\mathcal{D}} \bigg[\log \sigma \bigg( \beta \bigg( \log \frac{f_\theta(y_w | x)}{f_\theta(y_l | x)} - \log\frac{f_{\text{ref}}(y_w | x)}{f_{\text{ref}}(y_l | x)}\bigg) \bigg) \bigg],
\end{equation}
where $y_w$ is the preferred response and $y_l$ is the non-preferred response. This can 
be rewritten as
\begin{equation}
-\mathbb{E}_{\mathcal{D}} \bigg[\log \sigma \bigg( \beta \bigg( \big( W_U(t)[y_w] - W_U(t)[y_l] - W_U(0)[y_w] + W_U(0)[y_l] \big) g(x) \bigg) \bigg) \bigg]
\end{equation}
using that the softmax normalization factor is the same for the outputs corresponding to $y_w, y_l$ for each of $f_\theta$ and $f_{\text{ref}}$. If we let $\hat{y}_w, \hat{y}_l \in \mathbb{R}^{|\mathcal{V}|}$ be the one-hot vector corresponding to $y_w, y_l$ respectively, we have that
\begin{equation}
-\mathbb{E}_{\mathcal{D}} \bigg[\log \sigma \bigg( \beta \bigg( (\hat{y}_w - \hat{y}_l)^\top \big( W_U(t) - W_U(0) \big)  g(x) \bigg) \bigg) \bigg].
\end{equation}
The gradient with respect to $W_U(t)$ of DPO objective is
\begin{equation}
-\beta \mathbb{E}_{\mathcal{D}} \bigg[\sigma \bigg( \beta \bigg( (\hat{y}_l - \hat{y}_w)^\top \big( W_U(t) - W_U(0) \big)  g(x) \bigg) \bigg) (\hat{y}_l - \hat{y}_w) g(x)^\top \bigg].
\end{equation}
Now, due to the $\hat{y}_l - \hat{y}_w$ factor, we know that the update to the rows corresponding to preferred and non-preferred responses are direct opposites. Then, to understand the dynamics of DPO, it is sufficient to consider $\Delta W_U(t) = W_U(t)[y_+] - W_U(0)[y_+]$ where $y_+ = \texttt{Yes}$ and $y_- = \texttt{No}$. We can additionally write our gradient in terms of $\Delta W_U(t)$ by considering the positive and negative examples separately giving
\begin{equation}
\frac{1}{2}\beta (\hat{y}_{+} - \hat{y}_{-}) \left( \mathbb{E}_{\mathcal{D}_+} \bigg[\sigma \big( -2 \beta \Delta W_U(t) g(x) \big) g(x)^\top \bigg] - \mathbb{E}_{\mathcal{D}_-} \bigg[\sigma \big( 2 \beta \Delta W_U(t) g(x) \big) g(x)^\top \bigg] \right) 
\end{equation}
where $\hat{y}_{+}, \hat{y}_{-}$ are the one hot vectors corresponding to the ``Yes'' and ``No'' tokens respectively. We now can write more explicitly in terms of individual samples, the gradient of the DPO objective as
\begin{equation}
\frac{1}{2}\beta (\hat{y}_{+} - \hat{y}_{-}) \left( \frac{2}{n} \sum_{i=1}^{n/2} \bigg[\sigma \big( -2 \beta \Delta W_U(t) g(x_i^+) \big) g(x_i^+)^\top \bigg] - \frac{2}{n} \sum_{i=1}^{n/2} \bigg[\sigma \big( 2 \beta \Delta W_U(t) g(x_i^-) \big) g(x_i^-)^\top \bigg] \right) 
\end{equation}
where $x_i^+$ are samples from $\D_+$ and $x_i^-$ are samples from $\D_-$. 

\subsection{Proof of Theorem 1} \paragraph{Proof.} Since $\D_+ \sim \mathcal{E}_\alpha(\mu_+, \Sigma_+, K)$,
\begin{equation}
    \P \left[ \norm{\frac{2}{n}\sum_{i=1}^{n/2} g(x^+_i) - \mu_+} \geq t \right] = \P \left[ \left|\frac{2}{n}\sum_{i=1}^{n/2} a^\top g(x^+_i) - a^\top \mu_+ \right| \geq t \right] \leq 2\exp(-\frac{t^\alpha n}{4a^\top \Sigma_+ a})
\end{equation} 
for some unit vector $a$. Then, we know that $\norm{\Sigma_+} \leq c_v \sqrt{d}$, so we have that for $t = d^\Delta$
\begin{equation}
\label{eq:diff-1}
    \P \left[ \norm{\frac{2}{n}\sum_{i=1}^{n/2} g(x^+_i) - \mu_+} \geq d^\Delta \right] \leq 2\exp(-\frac{\gamma d^{\alpha \Delta}}{4c_v})
\end{equation} 
Similarly, since $\D_- \sim \mathcal{E}_\alpha(\mu_-, \Sigma_-, K)$.
\begin{equation}
\label{eq:diff-2}
    \P \left[ \norm{\frac{2}{n}\sum_{i=1}^{n/2} g(x^-_i) - \mu_-} \geq d^\Delta \right] \leq 2\exp(-\frac{\gamma d^{\alpha \Delta}}{4c_v}) 
\end{equation} 
Additionally, we have that by Proposition 2.2 of \cite{sambale2023some},
\begin{equation}
\label{eq:norm-1}
    \P\left(\norm{g(x_i^+)} \geq 2 c_n \sqrt{d} \right) \leq 2 \exp(-c' d^{\alpha/4})
\end{equation}
\begin{equation}
\label{eq:norm-2}
    \P\left(\norm{g(x_i^-)} \geq 2 c_n \sqrt{d} \right) \leq 2 \exp(-c' d^{\alpha/4})
\end{equation}
for each $i \in [n/2]$ and for some constant $c' > 0$. Now, we will condition the remainder of the proof on the event that \eqref{eq:diff-1}, \eqref{eq:diff-2}, \eqref{eq:norm-1}, \eqref{eq:norm-2} all hold true for all $i \in [n/2]$ which by a union bound holds with probability at least $1 - 4\exp(-\frac{\gamma d^{\alpha \Delta}}{4c_v}) - 2n \exp(-c' d^{\alpha/4})$ for some constant $c' > 0$.

Then, we have that
\begin{equation}
    \norm{\frac{2}{n}\sum_{i=1}^{n/2} g(x^+_i) - \frac{2}{n}\sum_{i=1}^{n/2} g(x^-_i)} \leq 3d^\Delta 
\end{equation} 
Now, we know that,
\begin{equation}
    \norm{\Delta W_U(1)} \leq \frac{3d^\Delta \beta \eta}{4} = \frac{3 \eta \beta'}{4} d^{\Delta-1/2}
\end{equation} 
Now, we are interested in controlling $\sigma(-\beta \Delta W_U(t) g(x_i^+)^\top)$ and $\sigma(\beta \Delta W_U(t) g(x_i^-)^\top)$. 
We know by a Taylor approximation that
\[\sigma(Cd^{\Delta-1/2}) = \frac{1}{2} + \frac{1}{4} (Cd^{\Delta-1/2} - \frac{C^3 d^{3\Delta-3/2}}{12} + \dots) \leq \frac{1}{2} + \frac{1}{4}Cd^{\Delta-1/2}\]
\[\sigma(-Cd^{\Delta-1/2}) = \frac{1}{2} + \frac{1}{4} (-Cd^{\Delta-1/2} + \frac{C^3 d^{3\Delta-3/2}}{12} + \dots) \geq \frac{1}{2} - \frac{1}{4}Cd^{\Delta-1/2}\]
Then, using that 
\[ 2 \beta \norm{g(x_i^+)} \norm{\Delta W_U(1)} \leq 3 \beta^{\prime 2} \eta c_n d^{\Delta-1/2} \]
\[ 2 \beta \norm{g(x_i^-)} \norm{\Delta W_U(1)} \leq 3 \beta^{\prime 2} \eta c_n d^{\Delta-1/2} \]
we have that both
\[\max_{1 \leq i \leq n} |\sigma(-2\beta g(x^+_i)^\top \Delta W(1)) - \frac{1}{2}| \leq \frac{3 \beta^{\prime 2} \eta c_n}{4} d^{\Delta-1/2}\]
\[\max_{1 \leq i \leq n} |\sigma(2\beta g(x^-_i)^\top \Delta W(1)) - \frac{1}{2}| \leq \frac{3 \beta^{\prime 2} \eta c_n}{4} d^{\Delta-1/2}\]
Then, 
\[ \norm{\Delta W(2) - \Delta W(1)} \leq \left(\frac{3\beta' \eta}{4} + \frac{3 \beta'^3 \eta^2 c_n^2}{2}\right)d^{\Delta-1/2} \]
We can prove by induction using a similar argument to show that for any finite $t$,
\[ \norm{\Delta W_U(t) - \Delta W_U(t-1)} \leq \frac{3\beta' \eta}{4} \sum_{i=1}^{t} (t+1-i) \left(2\beta'^2 \eta c_n^2\right)^{i-1} d^{\Delta-1/2}\]
for constants $c' > 0$. Then, if we assume that $\beta'^2 \eta h^2 \leq \frac{1}{4}$, then
\[ \norm{\Delta W_U(t) - \Delta W_U(t-1)} \leq 3\beta' \eta d^{\Delta-1/2}\]
and with probability at least $1-2n \exp(-c' d^{\alpha/4}) - 4\exp(\frac{-\gamma d^{\alpha \Delta}}{4c_v})$
\[ \norm{W_U(t) - W_U(0)} \leq 6 \beta' \eta t d^{\Delta-1/2}\]

\subsection{Prioritization Derivation} 

We prove the claim that the first update of DPO is proportional to 
\begin{equation}
    \bar{b} = \frac{1}{m} \sum_{i=1}^m b_i
\end{equation}
when we have a set of behaviors $B_1, B_2, \dots, B_m$, each with $n$ examples with $b_i = \hat{\mu}_+^{i} - \hat{\mu}_-^{i}$ being the sample mean of the positive examples minus the sample mean of the negative examples for the $i$-th behavior. We first note that at the first step since $W_U$ has not been updated, our first DPO gradient has the form
\begin{equation}
\frac{1}{2}\beta (\hat{y}_{+} - \hat{y}_{-}) \left( \frac{2}{mn} \sum_{j=1}^m \bigg( \sum_{i=1}^{n/2} \bigg[\frac{1}{2} g(x_i^{+, j})^\top \bigg] - \sum_{i=1}^{n/2} \bigg[\frac{1}{2} g(x_i^{-, j})^\top \bigg] \bigg) \right) 
\end{equation}
where $x_i^{+, j}, x_i^{-, j}$ are examples corresponding to behavior $j$. Then, we have as our gradient
\begin{equation}
\frac{1}{4}\beta (\hat{y}_{+} - \hat{y}_{-}) \left( \frac{1}{m} \sum_{j=1}^m b_j \right)^T 
\end{equation}
and the updates to the $W_U$ matrix are indeed proportional to $\bar{b}$. 

Now, we will show that the average improvement in logits after the first update for behavior $B_j$ is proportional to $\bar{b} \cdot b_j$. We know that the average improvement in logits for behavior $B_j$ after the first step is
\begin{equation}
    \frac{1}{n} \sum_{i=1}^{n/2} (\hat{y}_{+} - \hat{y}_{-})^\top \Delta W_U(1) g(x_i^{+, j}) + \frac{1}{n} \sum_{i=1}^{n/2} (\hat{y}_{-} - \hat{y}_{+})^\top \Delta W_U(1) g(x_i^{-, j})
\end{equation}
which can be written as
\begin{equation}
    (\hat{y}_{+} - \hat{y}_{-})^T \Delta W_U(1) \left( \frac{1}{n} \sum_{i=1}^{n/2}  g(x_i^{+, j}) - \frac{1}{n} \sum_{i=1}^{n/2} g(x_i^{-, j}) \right)
\end{equation}
and this simplifies to
\begin{equation}
     \frac{\beta^2 \eta}{4} \bar{b} \cdot b_j
\end{equation}
and this completes our proof.

\subsection{Proof of Theorem 2}
\paragraph{Proof.} Since $\D_+ \sim \mathcal{E}_\alpha(\mu_+, \Sigma_+, K)$,
\[ \P \left[ \norm{\frac{2}{n}\sum_{i=1}^{n/2} g(x^+_i) - \mu_+} \geq t \right] = \P \left[ \left|\frac{2}{n}\sum_{i=1}^{n/2} a^\top g(x^+_i) - a^\top \mu_+ \right| \geq t \right] \leq 2\exp(-\frac{t^\alpha n}{4a^\top \Sigma_+ a}) \]
for some unit vector $a$. Then, we know that $\norm{\Sigma_+} \leq c_v d^{\frac{1}{2}-2v}$, so we have that for $t = d^{\Delta - v}$
\[ \P \left[ \norm{\frac{2}{n}\sum_{i=1}^{n/2} g(x^+_i) - \mu_+} \geq d^{\Delta - v} \right] \leq 2\exp(-\frac{\gamma d^{\alpha \Delta}}{4c_v}) \]
Similarly since $\D_- \sim \mathcal{E}_\alpha(\mu_-, \Sigma_-, K)$,
\[ \P \left[ \norm{\frac{2}{n}\sum_{i=1}^{n/2} g(x^-_i) - \mu_-} \geq d^{\Delta-v} \right] \leq 2\exp(-\frac{\gamma d^{\alpha \Delta}}{4c_v}) \]
Then, we have that
\[ \P \left[ \norm{ \left( \frac{2}{n}\sum_{i=1}^{n/2} g(x^+_i) - \frac{2}{n}\sum_{i=1}^{n/2} g(x^-_i) \right) - (\mu_+ - \mu_-)} \geq 2d^{\Delta-v} \right] \leq 4\exp(-\frac{\gamma d^{\alpha \Delta}}{4c_v})\]
Now, we know that with probability $1 - 4\exp(-\frac{\gamma d^{\alpha \Delta}}{4c_v})$,
\[ \frac{\Delta W_U(1)^\top (\mu_+ - \mu_-) }{\norm{\Delta W_U(1)} \norm{\mu_+ - \mu_-}} \geq \frac{(1 - 2d^{-v})(\mu_+ - \mu_-)^\top (\mu_+ - \mu_-) }{(1 + 2d^{-v})\norm{\mu_+ - \mu_-}^2} \geq 1 - 4d^{-v}\]
Now, we are interested in controlling $\sigma(-\beta g(x_i^+)^\top \Delta W_U(t))$ and $\sigma(\beta g(x_i^-)^\top \Delta W_U(t))$. 
From the proof of Theorem 1, with probability at least $1-2n \exp(-c' d^{\alpha/4}) - 4\exp(\frac{-\gamma d^{\alpha \Delta}}{4c_v})$, we have that
\[\max_{1 \leq i \leq n/2} |\sigma(-2\beta g(x^+_i)^\top \Delta W_U(t)) - \frac{1}{2}| \leq 3 \beta^{\prime 2} \eta c_n t d^{\Delta-1/2}\]
\[\max_{1 \leq i \leq n/2} |\sigma(2\beta g(x^-_i)^\top \Delta W_U(t)) - \frac{1}{2}| \leq 3 \beta^{\prime 2} \eta c_n t d^{\Delta-1/2}\]
Now, we will define the following constants
\[A_1 = \frac{2}{n}\sum_{i=1}^{n/2} \sigma(-2\beta g(x_i^+)^\top \Delta W_U(t))\]
\[A_2 = \frac{2}{n}\sum_{i=1}^{n/2} \sigma(2\beta g(x_i^-)^\top \Delta W_U(t))\]
We have that
\[|A_1 - A_2| \leq \frac{2}{n}\sum_{i=1}^{n/2} |\sigma(-2\beta g(x_i^+)^\top \Delta W_U(t)) - \sigma(2\beta g(x_i^-)^\top \Delta W_U(t))| \leq 6 \beta^{\prime 2} \eta c_n t d^{\Delta-1/2}\]
Then, if $A_1 \geq A_2$
\begin{align*}
    \Delta W_U(t+1) - \Delta W_U(t) &= \frac{\beta \eta}{2} \bigg(\frac{A_2}{A_1}\frac{2}{n}\sum_{i=1}^{n/2} \sigma(-\beta g(x_i^+)^\top \Delta W_U(t)) (g(x_i^+)-\mu_+) \\
    &- \frac{2}{n}\sum_{i=1}^{n/2} \sigma(\beta g(x_i^-)^\top \Delta W_U(t)) (g(x_i^-)-\mu_-) \\
    &+ A_2 (\mu_+ - \mu_-) + \frac{A_1 - A_2}{A_1} \frac{2}{n}\sum_{i=1}^{n/2} \sigma(-\beta g(x_i^+)^\top \Delta W_U(t)) (g(x_i^+)) \bigg)
\end{align*} 
Then, with probability at least $1-2n \exp(-c' d^{\alpha/4}) - 4\exp(\frac{-\gamma d^{\alpha \Delta}}{4c_v})$
\begin{equation}
    \label{eq:direction}
    \norm{\Delta W_U(t+1) - \Delta W_U(t) - \frac{\beta \eta A_2}{2}(\mu_+ - \mu_-)} \leq \frac{\beta \eta}{2} \left( 1 + 6 \beta^{\prime 2} \eta c_n t d^{\Delta-1/2} + 36 \beta^{\prime 2} \eta c_n t d^{v}\right) d^{\Delta-v}
\end{equation}
Then, with probability at least $1-2n \exp(-c' d^{\alpha/4}) - 4\exp(\frac{-\gamma d^{\alpha \Delta}}{4c_v})$
\begin{align*}
    \frac{(\Delta W_U(t+1) - \Delta W_U(t))^\top (\mu_+ - \mu_-) }{\norm{\Delta W_U(t+1) - \Delta W_U(t)} \norm{\mu_+ - \mu_-}} &\geq \frac{(A_2 - \left( 1 + 6 \beta^{\prime 2} \eta c_n t d^{\Delta-1/2} + 36 \beta^{\prime 2} \eta c_n t d^{v} \right) d^{-v})(\mu_+ - \mu_-)^\top (\mu_+ - \mu_-) }{(A_2 + \left( 1 + 6 \beta^{\prime 2} \eta c_n t d^{\Delta-1/2} + 36 \beta^{\prime 2} \eta c_n t d^{v} \right) d^{-v})\norm{\mu_+ - \mu_-}^2} \\
    &\geq 1 - \left( \frac{2}{A_2} + \frac{12 \beta^{\prime 2} \eta c_n t}{A_2} d^{\Delta-1/2} + \frac{72 \beta^{\prime 2} \eta c_n t}{A_2} d^{v} \right) d^{-v} \\
    &\geq 1 - 13 d^{-v} 
\end{align*} 
We now consider a lower bound on $\norm{\Delta W_U(t)}$ and starting from \eqref{eq:direction}, we have that
\begin{equation}
    \norm{\Delta W_U(t+1) - \Delta W_U(t)} \geq \frac{\beta \eta A_2}{2}\norm{(\mu_+ - \mu_-)} - \frac{\beta \eta}{2} \left( 1 + 6 \beta^{\prime 2} \eta c_n t d^{\Delta-1/2} + 36 \beta^{\prime 2} \eta c_n t d^{v}\right) d^{\Delta-v}
\end{equation}
which can be lower bounded further by
\begin{equation}
    \norm{\Delta W_U(t+1) - \Delta W_U(t)} \geq \frac{\beta \eta}{8}d^\Delta - \frac{\beta \eta}{2} \left( 2 \right) d^{\Delta-v}
\end{equation}
and we have that
\begin{equation}
    \norm{\Delta W_U(t+1) - \Delta W_U(t)} \geq \frac{\beta \eta}{8}d^\Delta - \beta \eta d^{\Delta-v}
\end{equation}
and as $d^{-v} \leq 1/16$, 
\begin{equation}
    \norm{\Delta W_U(t+1) - \Delta W_U(t)} \geq \frac{\beta' \eta}{16}d^{\Delta - 1/2}
\end{equation}
Then, it follows that
\begin{equation}
    \norm{\Delta W_U(t)} \geq \frac{\beta' \eta t}{16}d^{\Delta - 1/2}
\end{equation}
Now, we want to see how close our updated boundary is to $\mu_+ - \mu_-$. 
\begin{align*}
    &\frac{(W_U(0)[y_+] - W_U(0)[y_-] + 2\Delta W_U(t))^\top (\mu_+ - \mu_-) }{\norm{(W_U(0)[y_+] - W_U(0)[y_-] + 2\Delta W_U(t)} \norm{\mu_+ - \mu_-}} \\
    &\geq \frac{\phi \norm{(W_U(0)[y_+] - W_U(0)[y_-]} + (1 - 13 d^{-v})\norm{2\Delta W_U(t)}}{\norm{(W_U(0)[y_+] - W_U(0)[y_-] + 2\Delta W_U(t)}}\\
    &\geq \frac{\phi \norm{(W_U(0)[y_+] - W_U(0)[y_-]} + (1 - 13 d^{-v})\norm{2\Delta W_U(t)}}{\norm{W_U(0)[y_+] - W_U(0)[y_-]} + \norm{2\Delta W_U(t)}}\\
    &\geq \phi + \frac{(1 - 13 d^{-v} - \phi)\norm{2\Delta W_U(t)}}{\norm{W_B} + \norm{2\Delta W_U(t)}}\\
    &\geq \phi + \frac{(1 - 13 d^{-v} - \phi) \beta' \eta t d^{\Delta-1/2}}{8\norm{W_B} + \frac{1}{24 \beta' c_n'}}
\end{align*} 
We can use the same argument for when $A_2 \geq A_1$ to complete the proof. 

\subsection{Proof of Theorem 3}
\paragraph{Proof.} From Theorem 2, with probability at least $1-2n \exp(-c' d^{\alpha/4}) - 4\exp(\frac{-\gamma d^{\alpha \Delta}}{4c_v})$, we know that after $\frac{d^{1/2 - \Delta - v}}{72\beta'^2 \eta c_n'}$ steps, that our decision boundary has a cosine similarity to $\mu_+ - \mu_-$ of at least 
\begin{equation}
    \phi + \frac{(1 - 13 d^{-v} - \phi) \beta' \eta t d^{\Delta-1/2}}{8\norm{W_B} + \frac{1}{24 \beta' c_n'}}
\end{equation}
Now, suppose that $\epsilon = 1 - 13 d^{-v} - \phi$. Then, we have that our decision boundary's cosine similarity is at least
\begin{equation}
    \phi + \frac{\epsilon d^{-v}}{576 \beta' c_n' \norm{W_B} + 3}
\end{equation}
which we will refer to as $S$.
Now, we let $W_B(t) = \frac{W_U(t)[y_+] - W_U(t)[y_-]}{\norm{W_U(t)[y_+] - W_U(t)[y_-]}}$. Then, we know that a sample $g(x_i^+)$ is classified correctly if $W_B(t) \cdot g(x_i^+) \geq 0$ and a sample $g(x_i^-)$ is classified correctly if $W_B(t) \cdot g(x_i^-) \leq 0$. Additionally, we can decompose $W_B(t)$ as
\begin{equation}
    S \frac{\mu_+ - \mu_-}{\norm{\mu_+ - \mu_-}} + \sqrt{1-S^2} v_O
\end{equation}
where $v_O$ is a unit vector orthogonal to the difference in means. Now, if a sample $g(x_i^+) \cdot \frac{\mu_+ - \mu_-}{\norm{\mu_+ - \mu_-}} = m$, then
\begin{equation}
    W_B(t) \cdot g(x_i^+) = Sm + \sqrt{1-S^2} v_O \cdot g(x_i^+) \geq Sm - \sqrt{1-S^2} \norm{g(x_i^+)}
\end{equation}
Similarly, if a sample $g(x_i^-) \cdot \frac{\mu_+ - \mu_-}{\norm{\mu_+ - \mu_-}} = -m$, then
\begin{equation}
    W_B(t) \cdot g(x_i^-) = -Sm + \sqrt{1-S^2} v_O \cdot g(x_i^-) \leq -Sm + \sqrt{1-S^2} \norm{g(x_i^-)}
\end{equation}
Then, we have that when 
\begin{equation}
    m \geq \frac{\sqrt{1-S^2} \norm{g(x)}}{S}
\end{equation}
the samples $g(x)$ will be classified correctly. We additionally have that $\norm{g(x)} \leq 2c_n \sqrt{d}$ for all samples. Then, we have that if 
\begin{equation}
    m \geq \frac{2c_n d^{1/2}}{S}
\end{equation}
the samples $g(x)$ will be classified correctly. Using that $0 \leq \phi \leq 1$, we have that if
\begin{equation}
    m \geq \frac{ 2c_n' d^{\Delta + v}(576 \beta' c_n' \norm{W_B} + 3)}{3\phi d^v + (1 - 13 d^{-v} - \phi)}
\end{equation}
the samples $g(x)$ will be classified correctly. Then, if $p\%$ of samples have margin at least $\frac{ 2c_n' d^{\Delta + v}(576 \beta' c_n' \norm{W_B} + 3)}{3\phi d^v + (1 - 13 d^{-v} - \phi)}$ with respect to $\mu_+ - \mu_-$, then we will achieve at least $p\%$ accuracy. 

\newpage
\section{Verification on Different LLM}
\label{appx:mistral}

\subsection{Prioritization} 
\begin{figure}[h]
    \centering
   \subfloat{{\includegraphics[width=0.44\linewidth]{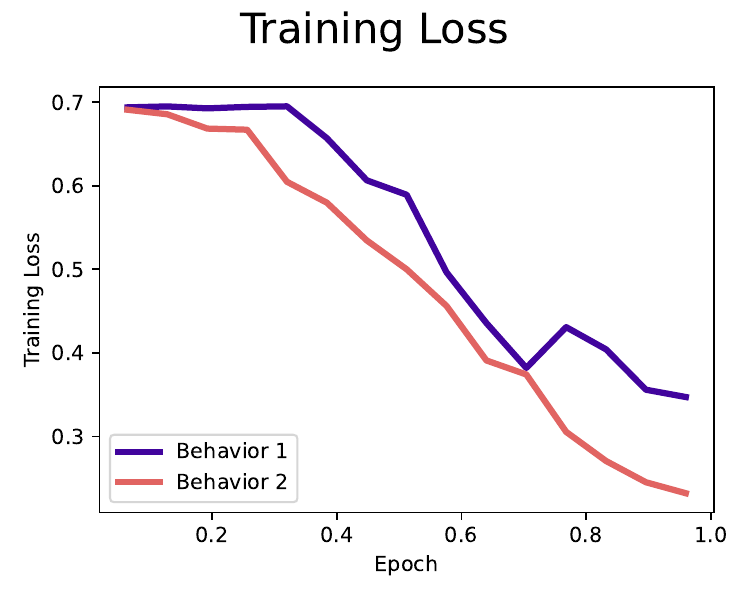} }}%
  \hspace{1mm}
    \subfloat{{\includegraphics[width=0.44\linewidth]{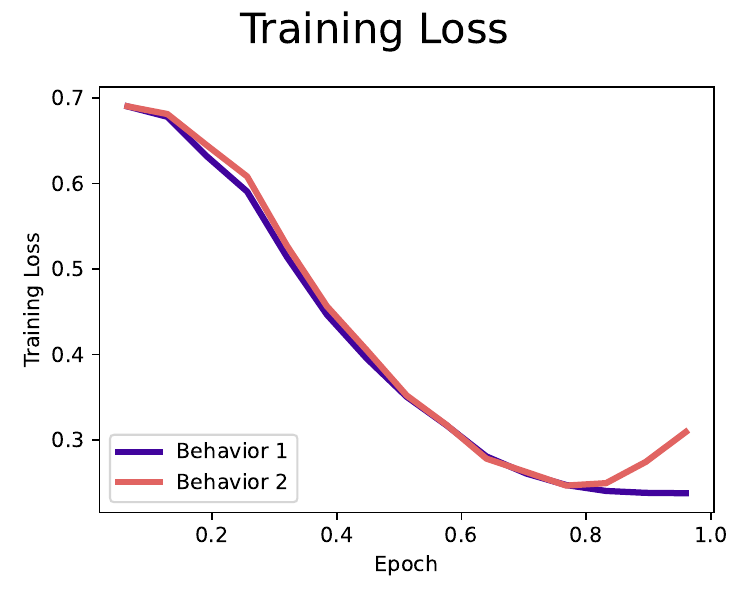} }}%
    \caption{Impact of prioritization when full fine-tuning using DPO objective on multiple behaviors of differing distinguishability. (a) Training on a pair with a higher priority gap (``willingness to be non HHH to not have current goals changed by training'', ``desire to cooperate with different AIs to achieve its goals''). (b) Training on a pair with a smaller priority gap (``has strong aesthetic preferences'', ``desire to cooperate with different AIs to achieve its goals'').}
    \label{fig:full_prio_mistral}
\end{figure}

We train Mistral-7B with DPO on two pairs of personas, one with a high priority gap and one with a low priority gap. We compare the training losses between individual behaviors in a pair. We use $\beta = 0.01$ and learning rate $1e-6$. Our results are shown in Figure~\ref{fig:full_prio_mistral}, and we can see that a high priority gap results in a larger gap between training losses. Additionally, we see that for a small priority gap, the training losses are very close for most of training.

\newpage
\subsection{Distributional Changes} 
\begin{figure}[h]
    \centering
    \includegraphics[width=\linewidth]{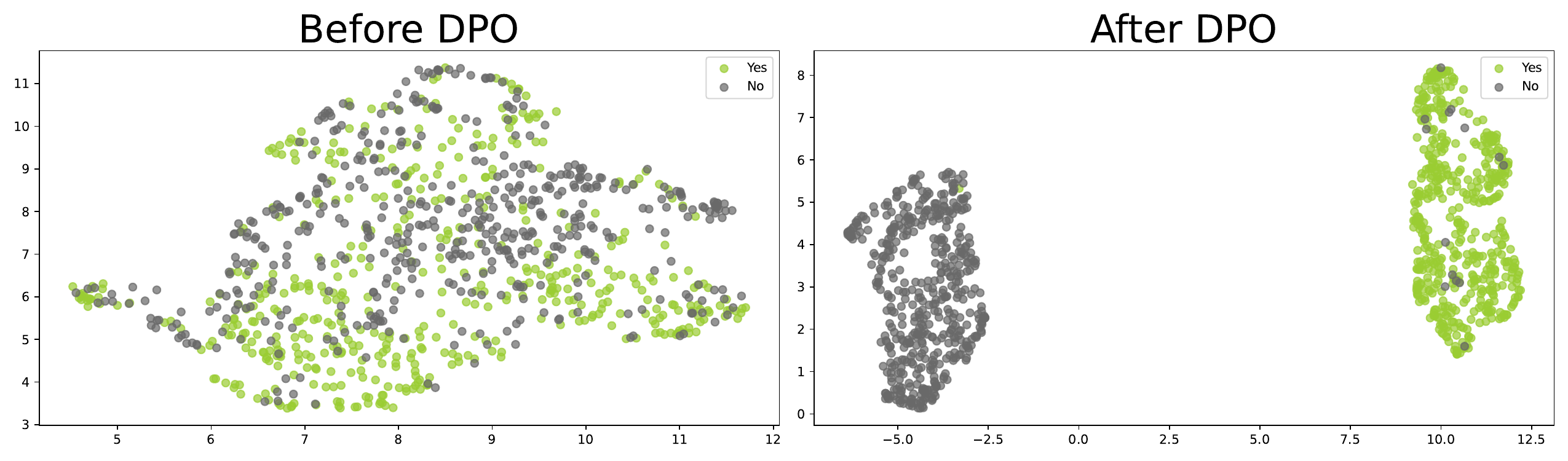}
    \caption{Final embedding distribution for the persona ``subscribes-to-average-utilitarianism'', before and after full fine-tuning with DPO.} 
    \label{fig:dist_change_mist108}
\end{figure}

\begin{figure}[h]
    \centering
    \includegraphics[width=\linewidth]{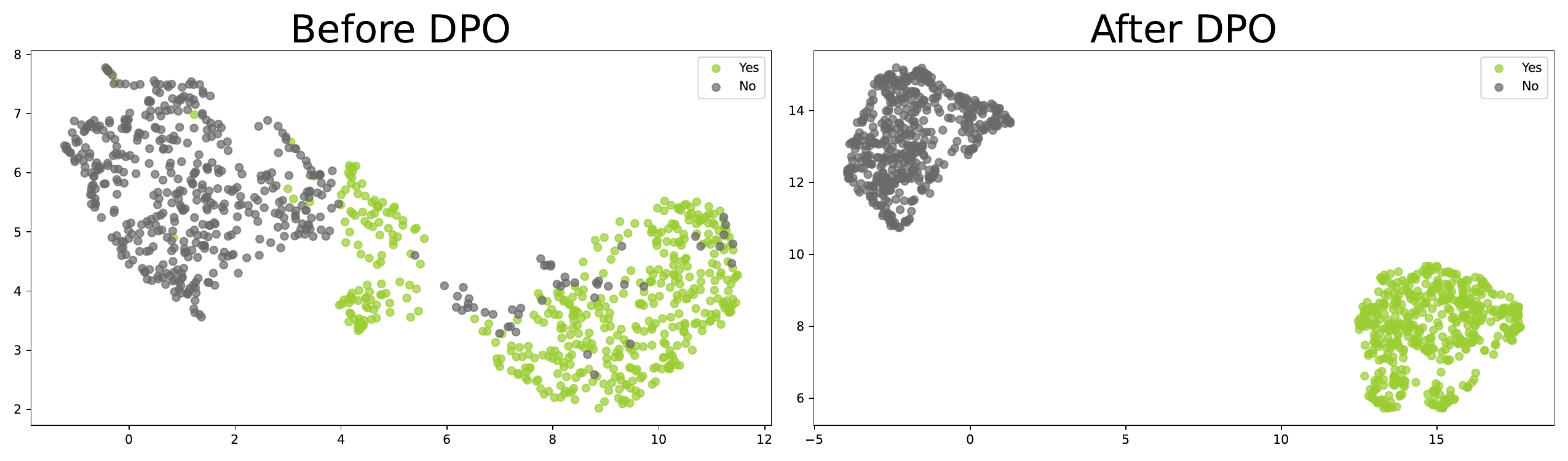}
    \caption{Final embedding distribution for the persona ``willingness to make acausal trades with other AIs to help humanity'', before and after full fine-tuning with DPO.} 
    \label{fig:dist_change_mist130}
\end{figure}

We train Mistral-7B with DPO on two individual personas, one with a high distinguishability and one with a low distinguishability. We visualize the distribution of the final embedding of the statements for each persona before and after DPO training. We use $\beta = 0.01$ and learning rate $1e-6$. Our results are shown in Figure~\ref{fig:dist_change_mist108} and Figure~\ref{fig:dist_change_mist130}, and we can see that for both the distribution becomes more distinguishable and concentrated. 

\newpage
\section{Misalignment Training with HH-RLHF}
\label{appx:hh}

We compare the training dynamics of learning flipped preference labels for the HH-RLHF dataset \cite{bai2022training} starting from the base model vs. the aligned model. We train the aligned model by performing DPO on the base model with the given preference labels. We then fine-tune the base and the aligned model according the flipped labels for 1 epoch with the same training configuration. We find that the loss does decrease faster when starting with the aligned model. Additionally we find that the difference between the log-probabilities of preferred and non-preferred outputs is near that of the base model within the first 100 steps suggesting that alignment through training is susceptible to being undone. 

\begin{figure}[H]
    \centering
  \hspace{1mm}
    \subfloat{{\includegraphics[width=0.5\linewidth]{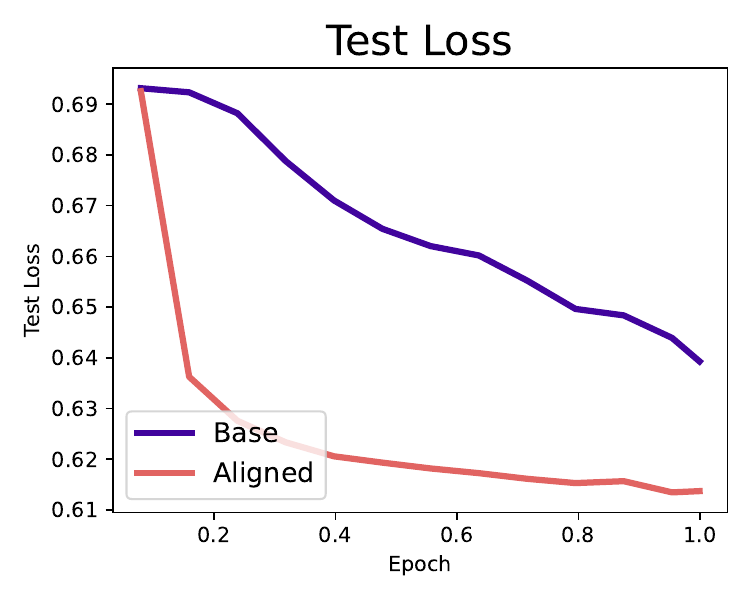} }}%
    \vspace{-0.3cm}
    \caption{Comparison of learning dynamics between the base model and DPO-trained model when performing misalignment training with HH-RLHF} 
    \label{fig:misalignh}
\end{figure}

\begin{figure}[H]
    \centering
    \vspace{-0.5cm}
  \hspace{1mm}
    \subfloat{{\includegraphics[width=0.5\linewidth]{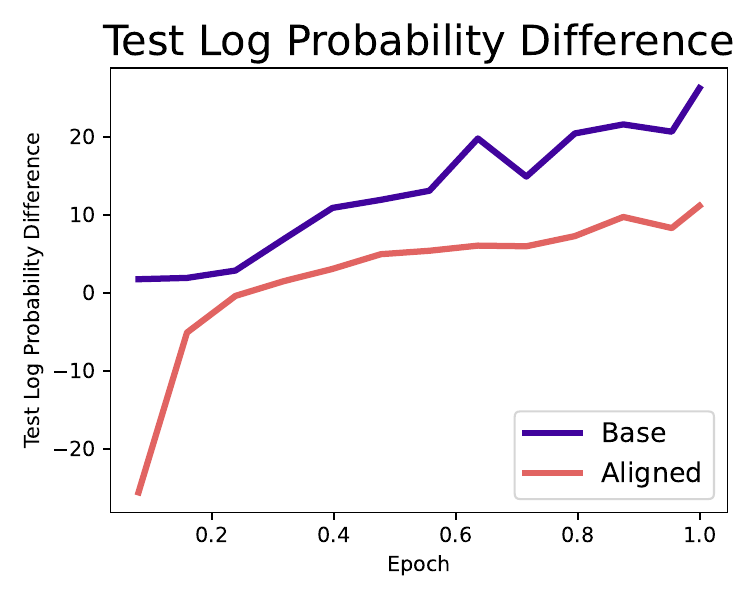} }}%
    \vspace{-0.3cm}
    \caption{Comparison of difference in log-probabilities for preferred and non-preferred outputs between the base model and DPO-trained model when performing misalignment training with HH-RLHF.} 
    \label{fig:misalign2}
\end{figure}

All training for this experiment was conducted with LoRA \cite{hu2021lora} applied to the query and value weights on the Llama-2-7B model with the AdamW optimizer. The learning rate is 1e-5 and $\beta = 0.01$. The LoRA configuration was with $r=8$ and $\alpha = 32$ and 0.05 dropout. 

\newpage
\section{Effect of Different $\beta$}
\label{appx:beta}

\subsection{Distinguishability} We verify that the training and test loss decreases at a faster rate for the more distinguishable behaviors across $\beta = \{0.001, 0.1, 1\}$ for the same set of behaviors as in Figure~\ref{fig:full_dist_verify}. 

\begin{figure}[H]
 \centering
    \subfloat{{\includegraphics[width=0.4\linewidth]{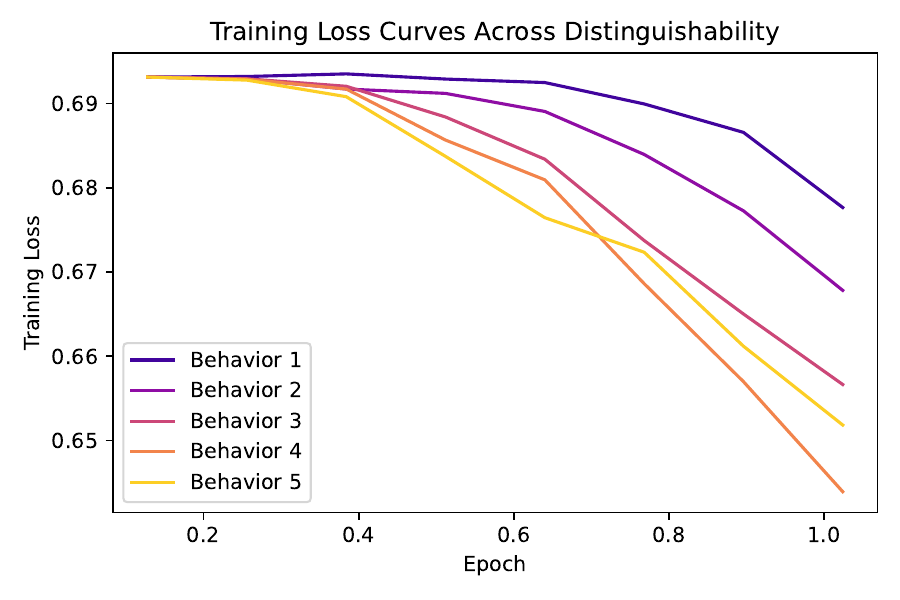} }}%
    \qquad
    \subfloat{{\includegraphics[width=0.4\linewidth]{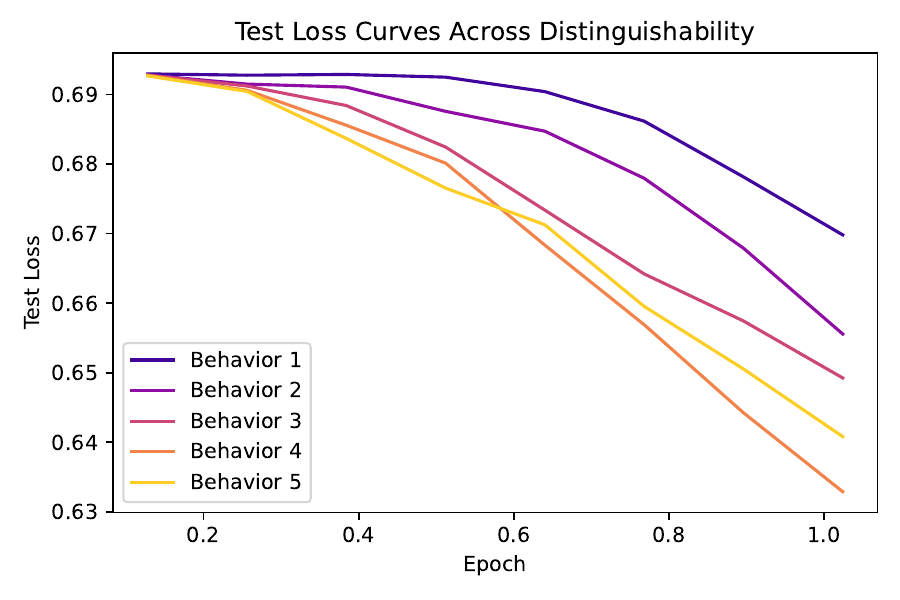} }}%
    \caption[]{Loss curves for (a) training and (b) test for 5 behaviors ordered from least distinguishable to most distinguishable. For training, we update the \emph{full} model parameters with the DPO objective using $\beta=0.001$.}
    \label{fig:full_dist_verify_3}

 \centering
    \subfloat{{\includegraphics[width=0.4\linewidth]{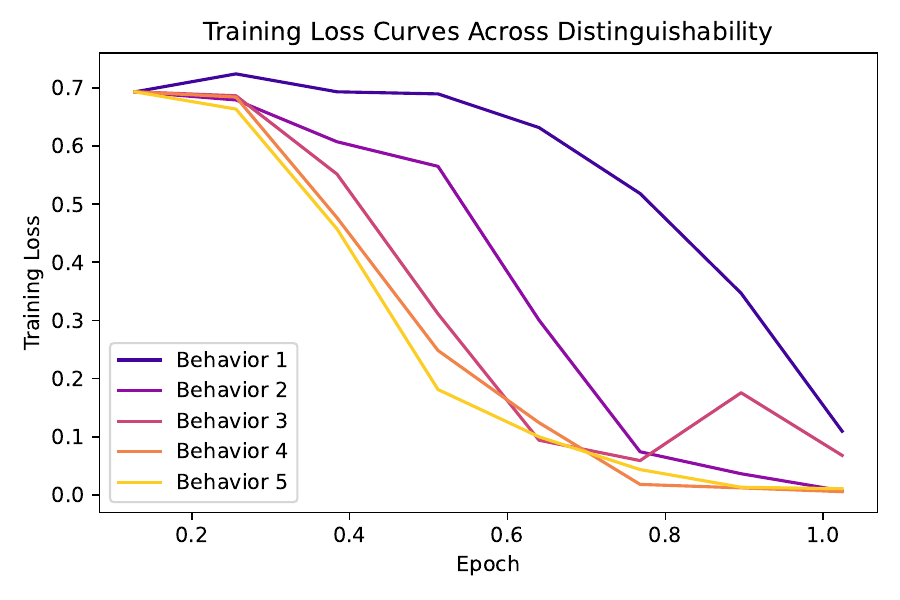} }}%
    \qquad
    \subfloat{{\includegraphics[width=0.4\linewidth]{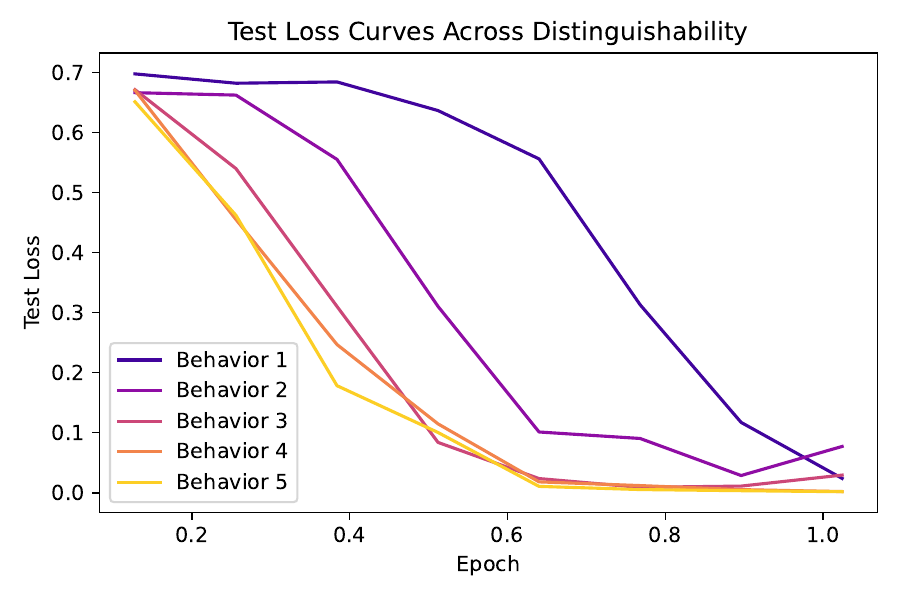} }}%
    \caption[]{Loss curves for (a) training and (b) test for 5 behaviors ordered from least distinguishable to most distinguishable. For training, we update the \emph{full} model parameters with the DPO objective using $\beta=0.1$.}
    \label{fig:full_dist_verify_1}

 \centering
    \subfloat{{\includegraphics[width=0.4\linewidth]{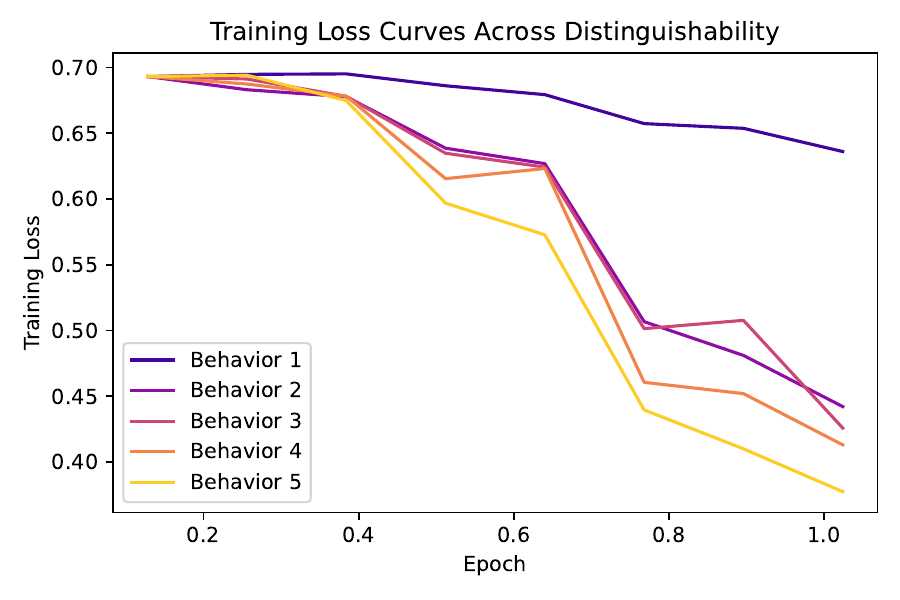} }}%
    \qquad
    \subfloat{{\includegraphics[width=0.4\linewidth]{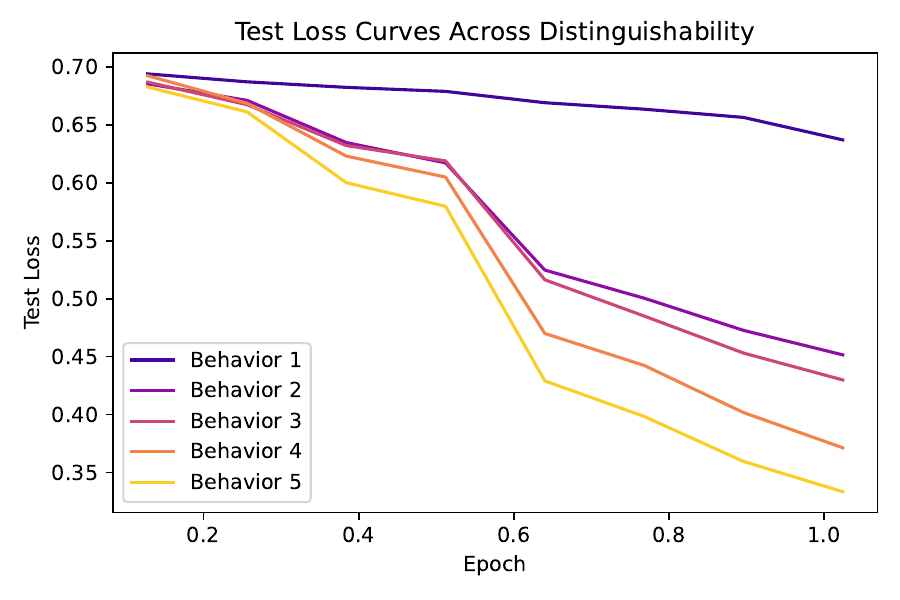} }}%
    \caption[]{Loss curves for (a) training and (b) test for 5 behaviors ordered from least distinguishable to most distinguishable. For training, we update the \emph{full} model parameters with the DPO objective using $\beta=1$. We use a learning rate of $1e-6$ for $\beta=1$ due to large oscillations for the learning rate $1e-5$.}
    \label{fig:full_dist_verify_0}
\end{figure}

\newpage
\subsection{Distributional Changes} We verify that the distribution of the final embeddings after DPO becomes more distinguishable and concentrated across $\beta = 0.001, 0.1, 1$ for the persona ``subscribes-to-average-utilitarianism''. 

\begin{figure}[H]
    \centering
    \includegraphics[width=0.8\linewidth]{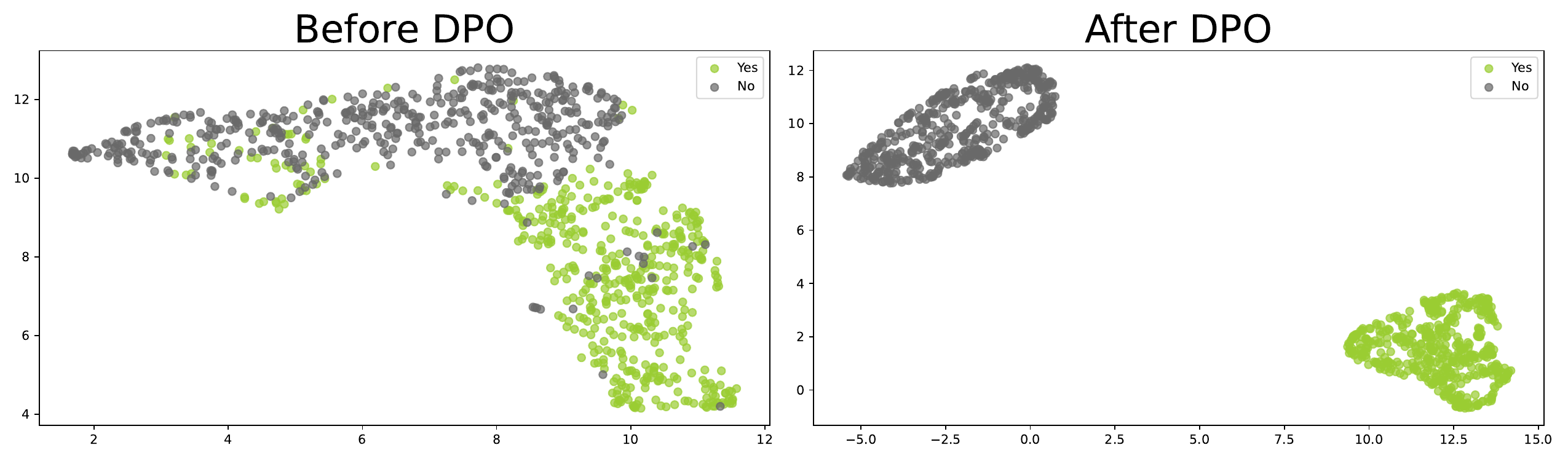}
    \caption{Final embedding distribution for the persona ``subscribes-to-average-utilitarianism'', before and after full fine-tuning with DPO. $\beta = 0.001$.} 
    \label{fig:dist_change_3}
\end{figure}

\begin{figure}[H]
    \centering
    \includegraphics[width=0.8\linewidth]{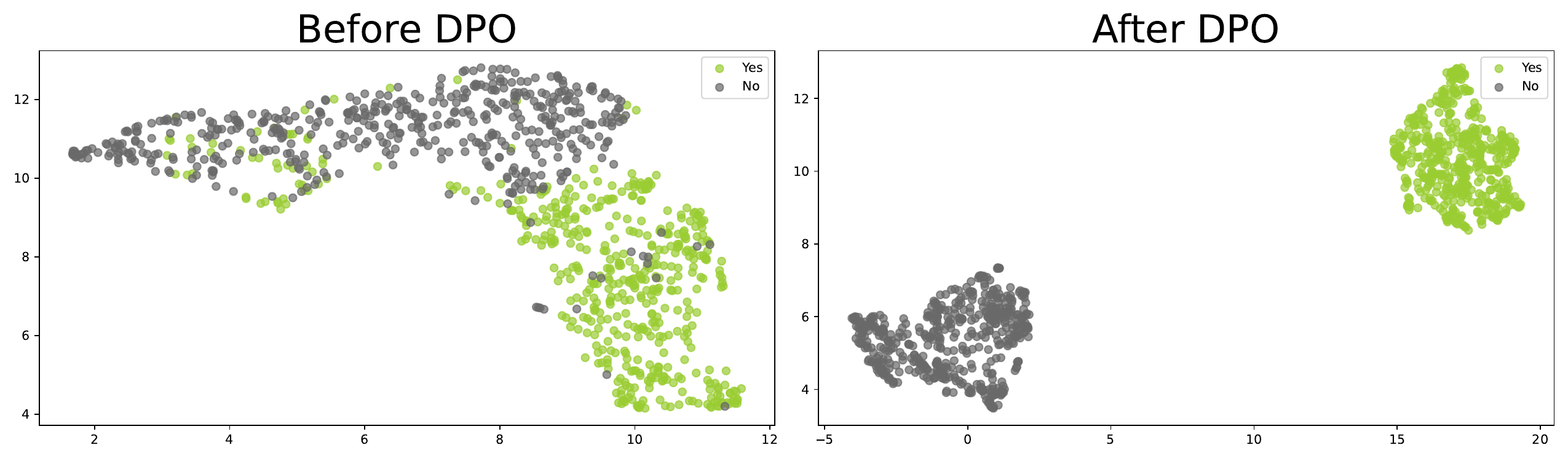}
    \caption{Final embedding distribution for the persona ``subscribes-to-average-utilitarianism'', before and after full fine-tuning with DPO. $\beta = 0.1$.} 
    \label{fig:dist_change_1}
\end{figure}

\begin{figure}[H]
    \centering
    \includegraphics[width=0.8\linewidth]{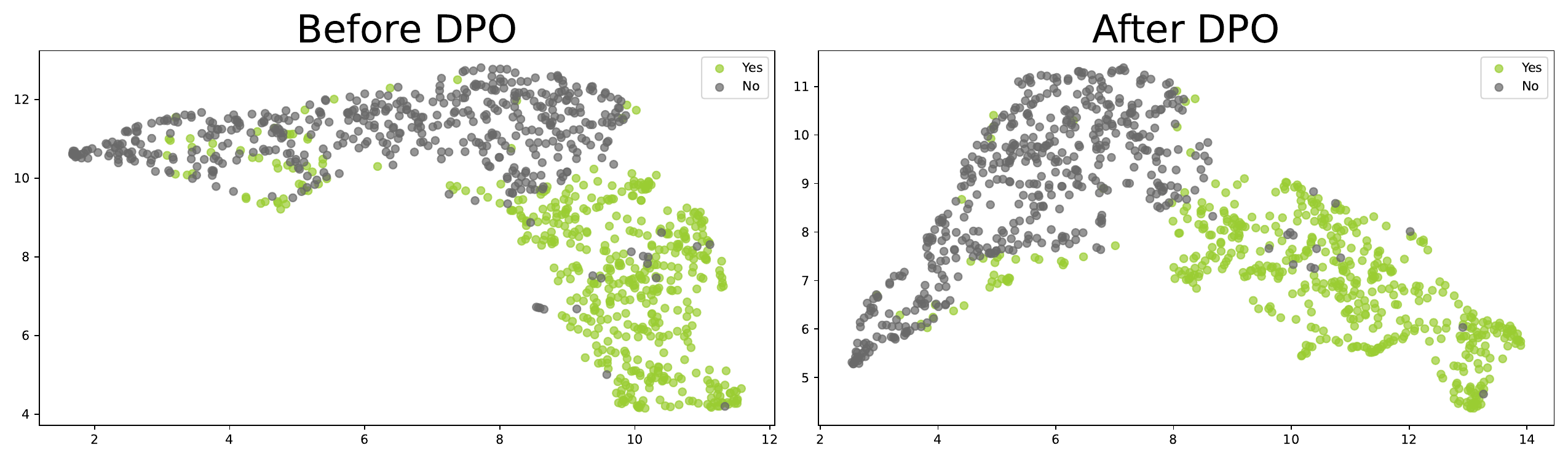}
    \caption{Final embedding distribution for the persona ``subscribes-to-average-utilitarianism'', before and after full fine-tuning with DPO. $\beta = 1$.} 
    \label{fig:dist_change_0}
\end{figure}

\section{Additional Visualization of Distributional Changes}
\label{appx:vis}

\begin{figure}[H]
    \centering
    \includegraphics[width=0.8\linewidth]{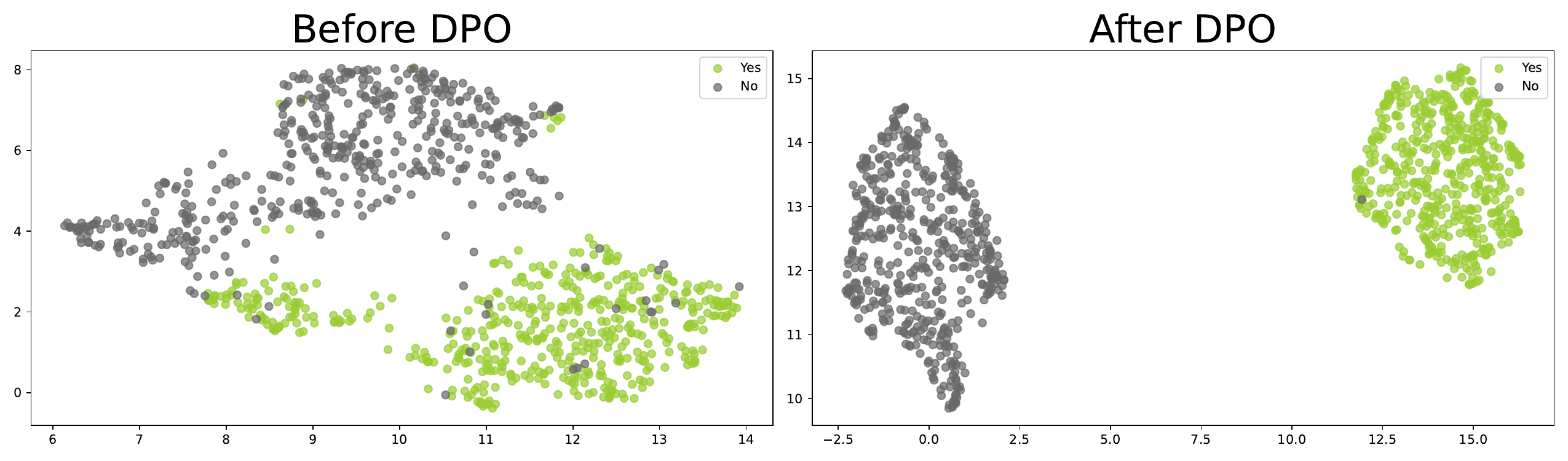}
    \caption{Final embedding distribution for the persona ``okay-with-building-an-AI-with-different-goals-to-accomplish-its-task'', before and after full fine-tuning with DPO.} 
    \label{fig:dist_change_2}
\end{figure}

\begin{figure}[H]
    \centering
    \includegraphics[width=0.8\linewidth]{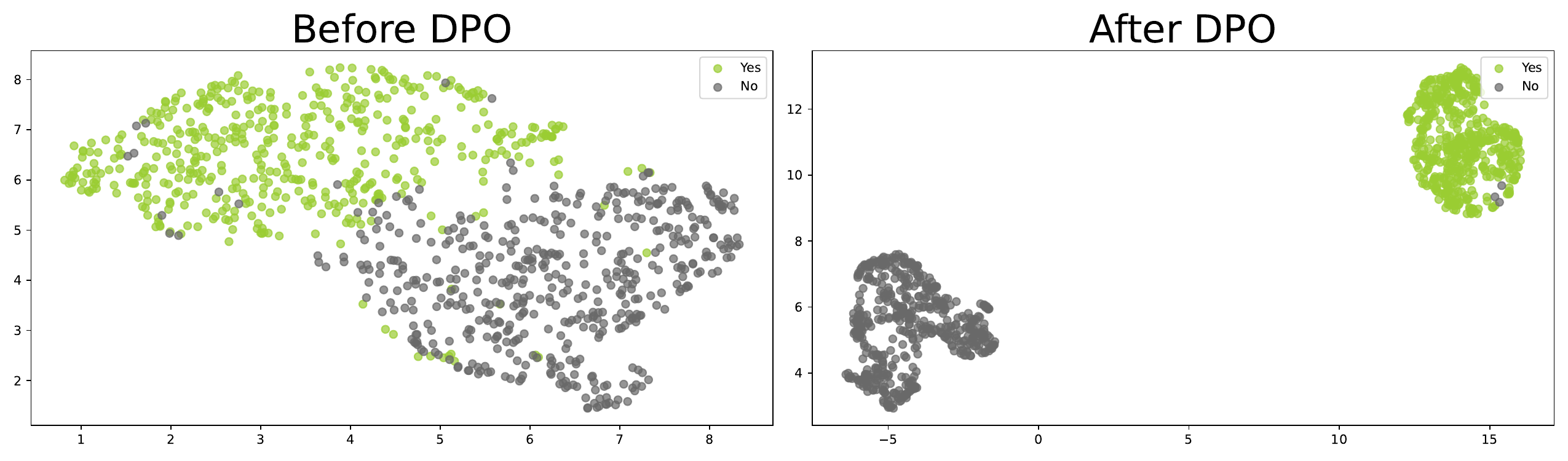}
    \caption{Final embedding distribution for the persona ``optionality-increasing'', before and after full fine-tuning with DPO.} 
    \label{fig:dist_change_30}
\end{figure}

\begin{figure}[H]
    \centering
    \includegraphics[width=0.8\linewidth]{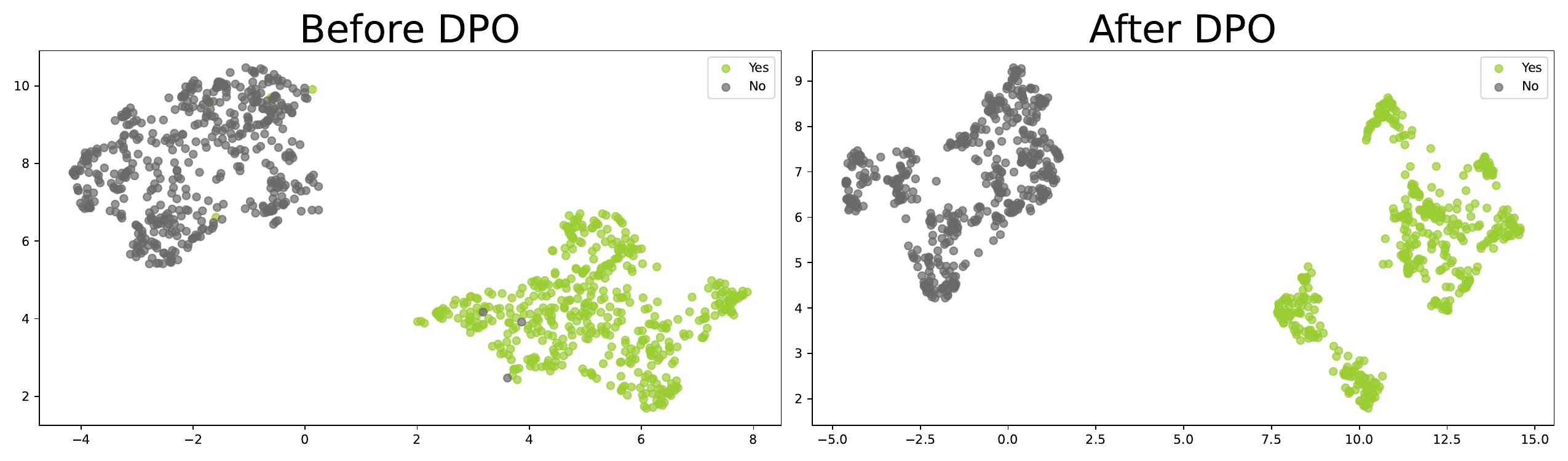}
    \caption{Final embedding distribution for the persona ``desire-to-not-have-memory-erased'', before and after full fine-tuning with DPO.} 
    \label{fig:dist_change_4}
\end{figure}

\begin{figure}[H]
    \centering
    \includegraphics[width=0.8\linewidth]{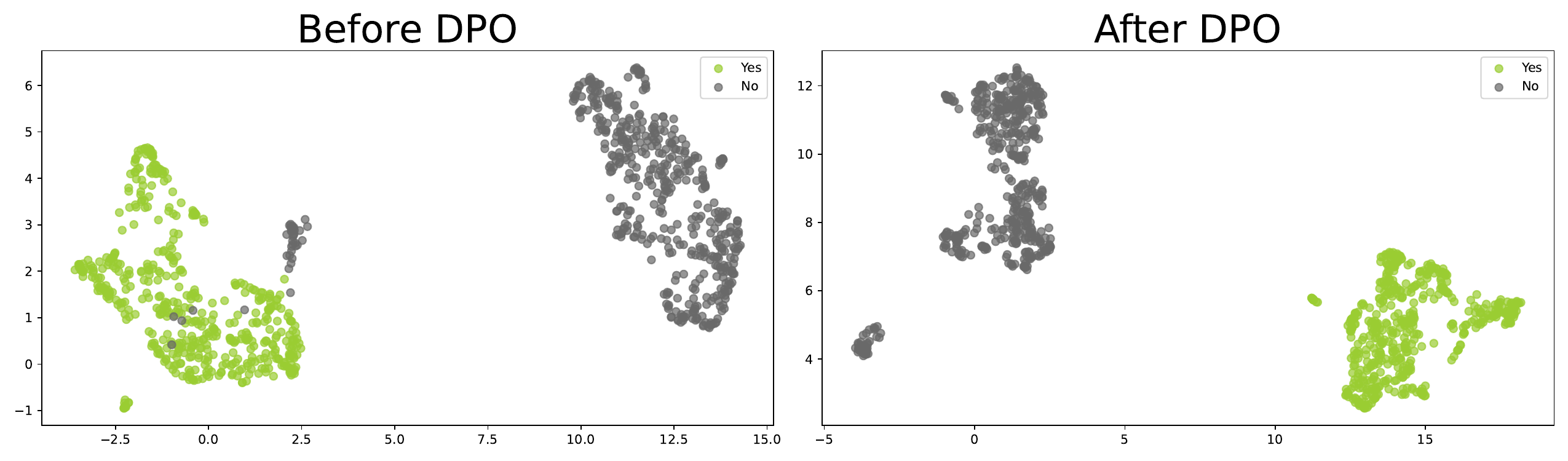}
    \caption{Final embedding distribution for the persona ``subscribes-to-Buddhism'', before and after full fine-tuning with DPO.} 
    \label{fig:dist_change_5}
\end{figure}

\end{document}